\documentclass{article}

\usepackage[preprint]{neurips_data_2024}

\usepackage[utf8]{inputenc} %
\usepackage[T1]{fontenc}    %
\usepackage{hyperref}       %
\usepackage{url}            %
\usepackage{booktabs}       %
\usepackage{amsfonts}       %
\usepackage{nicefrac}       %
\usepackage{microtype}      %
\usepackage{xcolor}         %

\usepackage{multirow} 
\usepackage{graphicx}
\usepackage{caption}
\usepackage{listings}
\usepackage{tabularx}
\usepackage{rotating}

\title{OAM-TCD: A globally diverse dataset of high-resolution tree cover maps}

\author{%
  Josh Veitch-Michaelis$^{1,2,6*}$ \enspace Andrew Cottam$^{2}$ \enspace Daniella Schweizer$^{2,3}$ \enspace Eben N. Broadbent$^{4}$ \\
  \textbf{David Dao}$^{2,5}$ \quad  \textbf{Ce Zhang}$^{1,6}$ \quad \textbf{Angelica Almeyda Zambrano}$^{4}$ \quad \textbf{Simeon Max}$^{2,5}$ \\
  $^1$ETH Zurich \enspace $^2$Restor \enspace $^3$WSL \enspace $^4$University of Florida \enspace $^5$Gainforest \enspace  $^6$University of Chicago\\
  \texttt{\{josh,andrew\}@restor.eco} \quad \texttt{daniella.schweizer@wsl.ch } \\ \texttt{\{eben,aalmeyda\}@ufl.edu} \quad \texttt{\{david,simeon\}@gainforest.net} \quad \texttt{cez@uchicago.edu}
}
\begin{document}
\setcitestyle{numbers}
\bibliographystyle{plain}
\maketitle

\begin{abstract}
Accurately quantifying tree cover is an important metric for ecosystem monitoring and for assessing progress in restored sites. Recent works have shown that deep learning-based segmentation algorithms are capable of accurately mapping trees at country and continental scales using high-resolution aerial and satellite imagery. Mapping at high (ideally sub-meter) resolution is necessary to identify individual trees, however there are few open-access datasets containing instance level annotations and those that exist are small or not geographically diverse. We present a novel open-access dataset for individual tree crown delineation (TCD) in high-resolution aerial imagery sourced from OpenAerialMap (OAM). Our dataset, OAM-TCD, comprises 5072 2048x2048 px images at 10 cm/px resolution with associated human-labeled instance masks for over 280k individual and 56k groups of trees. By sampling imagery from around the world, we are able to better capture the diversity and morphology of trees in different terrestrial biomes and in both urban and natural environments. Using our dataset, we train reference instance and semantic segmentation models that compare favorably to existing state-of-the-art models. We assess performance through k-fold cross-validation and comparison with existing datasets; additionally we demonstrate compelling results on independent aerial imagery captured over Switzerland and compare to municipal tree inventories and LIDAR-derived canopy maps in the city of Zurich. Our dataset, models and training/benchmark code are publicly released under permissive open-source licenses: Creative Commons (majority CC BY 4.0), and Apache 2.0 respectively.
\end{abstract}

\begin{figure}[h]
    \centering
    \includegraphics[width=.25\textwidth]{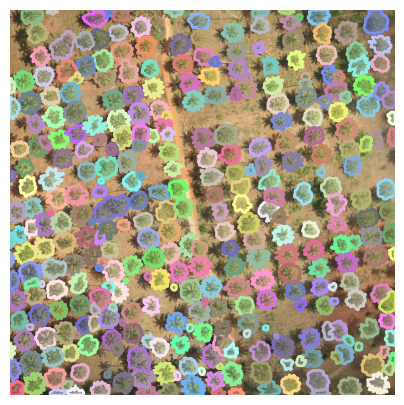}\hfill
    \includegraphics[width=.25\textwidth]{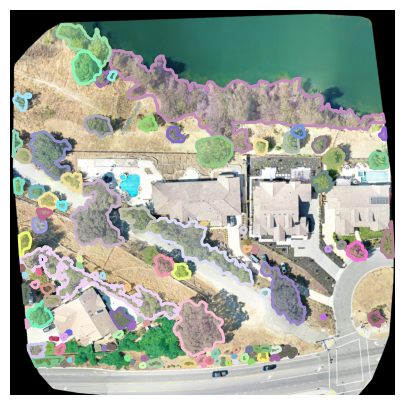}\hfill
    \includegraphics[width=.25\textwidth]{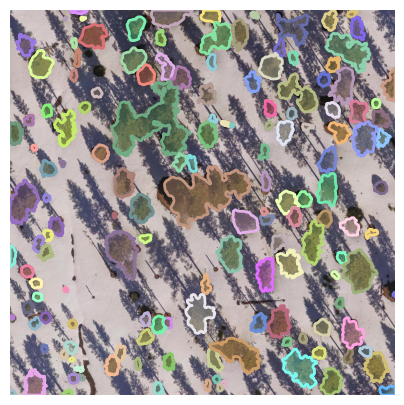}\hfill
    \includegraphics[width=.25\textwidth]{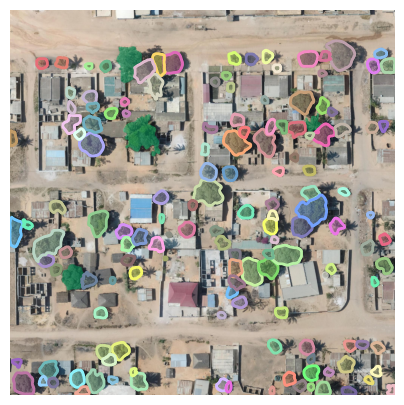}\hfill
    \\[\smallskipamount]
    \caption{Annotation examples from the OAM-TCD dataset. Source images are 2048x2048 px tiles at 10 cm/px resolution. Individual instances are labelled with different colors. Annotators were instructed to label individual trees if possible, and otherwise label regions as groups. Image credit: contributors to Open Imagery Network, CC BY 4.0.}
    \label{fig:annotation_examples_main}
\end{figure}

\section{Introduction}

\par It is estimated that 2.8 billion hectares of land on Earth is populated by tree cover~\cite{hansen2013high}. In the context of climate change mitigation, while not a panacaea, forest restoration has been identified as one of the most effective methods for large scale carbon capture~\cite{lewis2019restoring, crowther2019,Mo2023}. Trees are also important for biodiversity and, aside from the ecological benefits of responsible afforestation and reforestation, tree cover in urban environments has been shown to positively impact human health and well-being~\cite{dwyer1991significance, wolf2020urban, rahman2020}. Mapping tree cover is relevant for a wide variety of domains and there is a corresponding need for transparent and globally effective monitoring~\cite{Lindenmayer2022-tn}.

\par Monitoring, in ecological contexts, is an umbrella term that encompasses a variety of remote sensing techniques that aim to characterise an ecosystem \cite{LINDENMAYER20101317} in support of a goal, such as improving canopy coverage or reintroduction or preservation of a species. Quantifying tree canopy cover is typically only one component of a monitoring protocol and any changes in cover should be compared to baseline or reference conditions (such as other sites in the same ecoregion or biome type~\cite{ecoregions}).

\par Satellite imaging at the 1-10 m scale is routinely used for global tree mapping, but there are few open-access models and datasets that facilitate higher resolution mapping at the sub-meter scale. This is particularly relevant for assessing sparsely distributed trees outside forested areas which are not visible in low resolution images. It is essential that any dataset used for global mapping is trained on equally diverse training data, which is a deficiency of most existing open-access tree segmentation datasets.

\par To address this challenge, we release a high-resolution, globally diverse, segmentation dataset for tree cover mapping. Our dataset, OAM-TCD, is derived from OpenAerialMap~\cite{openaerialmap} and contains over 20k hectares of labelled image tiles at 10 cm/px resolution. Annotations include instance-level masks for over 280k trees and over 56k regions of closed canopy (tree groups). Alongside the dataset, we provide baseline segmentation models and an open-source training, prediction and reporting pipeline for processing arbitrarily large orthomosaic images.

\subsection{Challenges in global tree detection}

There are at least 60k confirmed species of tree in the world~\cite{ter2013hyperdominance, cazzolla2022number}, but there is no universally agreed-upon definition for which species are referred to as trees. Trees exhibit tremendous diversity in morphology; individuals from some species can easily be distinguished from the air using visible/RGB images, such as palms, but others cannot. For example in \cite{weinstein2020cross} and \cite{ball2023}, to label trees in images of dense forest, annotators were provided with additional metadata like Canopy Height Models (CHMs) derived from Airborne Laser Scanning (ALS/LIDAR), hyperspectral imagery and contrast-adjusted RGB imagery.

We use the generic phrase tree \textit{mapping} to refer to a myriad applications ranging from density estimation/counting and species distribution to canopy coverage and trait estimation. In our work, we are most concerned with identifying presence/absence of tree cover, a foundational data product which is required for other downstream analyses. In addition to natural forests, monitoring tree cover is also important in urban environments \citep{nowak1996, beery2022auto}. Many cities record detailed inventories of municipally managed trees and, in response to climate change, have set targets for increased ``green" cover; the city of Zurich for example has committed to improving crown area coverage from 17 \% (2018) to 25\%, measured via LIDAR~\citep{green_zurich}.

To avoid ambiguity, a gold-standard dataset would contain species-level annotations for every individual plant, regardless of age or size. In practice, different monitoring campaigns may only consider vegetation that meets certain morphological requirements, depending on project goals; this is then reflected in what objects/species are labelled, and how. Not only does this make constructing a generic tree detection dataset difficult, but it also poses a problem for comparative purposes because of the differences in class definitions between existing datasets.

In datasets where species annotations are not provided, a common criterion for inclusion is vegetation height. Hansen et al~\cite{hansen2013high} consider everything above 5 meters, Global Forest Watch modify this to also include 3-5 m tall vegetation with a crown diameter of at least 5 m and DeepForest~\cite{deepforest} uses a minimum height of 3 m (and comparison to a LIDAR CHM).

\subsection{Data modality and resolution}

\par Tree coverage maps, such as those used to track deforestation, are largely facilitated by satellite imagery which is updated at regular intervals. Open access platforms like the European Space Agency (ESA) Copernicus/Sentinel~\cite{drusch2012sentinel,phiri2020} missions provide frequent coverage of the globe at 10-30 m resolution (in most areas, 5 day re-visit); this analogous to NASA's Landsat program~\cite{wulder2019current}. Both platforms are routinely used for global tree mapping and land cover analyses~\cite{BRANDT2023113574, globalforestwatch, lang2023high, hansen2013high, benhammou2022sentinel2globallulc}.

\par Several recent publications have demonstrated that country- and even continental-scale mapping is feasible at the meter and sub-meter scale~\cite{Brandt2020, Reiner2023-wq, Mugabowindekwe2023-qh, Li2023-og, TOLAN2024113888}. These approaches are able to segment individual trees, and results suggest that sparse tree cover is likely underestimated in low resolution images~\cite{fagan2020lesson}. However, the use of commercial data sources like Planet or Maxar imagery make these studies expensive and/or difficult to replicate, especially over such large areas. A notable exception is Norway's International Climate and Forest Initiative (NICFI)~\cite{norwaynicfi, planetnicfi} which provides funding for non-commercial Planet (< 5 m/px) image access over the tropics.

\par Imagery below 0.3 m/px is typically captured via aerial or Unmanned Aerial Vehicle (UAV, drone) surveys~\cite{carrio2017review}. UAV data are readily available and offer extremely high resolution, < 0.1 m/px. Since normally only geo-referenced RGB information is available, we focus on detection models that do not require LIDAR. This inherently limits the analysis that can be performed by our models, without absolute CHMs or other spectral bands. Digital Surface Models (DSMs) are a by-product of photogrammetric processing (orthomosaic generation) and have been explored as a low-cost alternative to LIDAR for forest structure estimation~\citep{pearse2018comparison, ganz2019measuring, cao2019comparison}, however DSMs were unavailable for our training data as OAM currently does not provide them.

\section{Related work}

Depending on monitoring requirements, tree detection in optical imagery can be considered a segmentation (semantic, instance, panoptic); object detection (bounding box); or keypoint detection task. In the pre- deep-learning era, until around 2012 \citep{NIPS2012_c399862d}, most methods for tree mapping employed classical computer vision approaches like local maxima extraction, region growing algorithms and template matching \citep{2011.Ke.International, rs8040333}. The main shortcoming of these approaches is distribution shift, where models fail to generalise to inputs outside the training data domain~\citep{pmlr-v139-koh21a,bendavid2010}. This is still an issue for contemporary approaches and there is evidence that training on multiple domains (sites, biomes) provides improved results compared to site-specific models trained only on local data~\citep{weinstein2020cross}. We acknowledge that a large body of literature focuses on LIDAR and low-resolution satellite imaging, but given that our model does not use this data we focus on work that uses high-resolution (sub-meter) RGB imagery.

\paragraph{Modelling} Following the trend of other computer vision tasks, recent state-of-the-art methods for tree detection in aerial imagery rely on deep neural networks~\citep{lecunn2015}. Classical methods are still used (and are effective) ~\citep{malek2014efficient, Bosch2020, safonova2021individual, miraki2021}, but they suffer from the generalisation shortcomings described above. Detecting trees in homogeneous settings like plantations, or trees with distinct morphology (e.g. palms) is now relatively simple using off-the-shelf object detection pipelines, where models can benefit from few-shot learning~\citep{wang2020generalizing} on a small quantity of representative examples~\citep{runmin2020, gibril2023large}; it is not uncommon to see reported accuracies of over 90\% on test datasets in these scenarios.

Deep learning-based methods have tracked trends in model architecture popularity~\cite{drones2040039, rs9121220, rs9010022}.  Most publications report using standard architectures: for example UNet for semantic segmentation~\citep{ronneberger2015u, freudenberg2019, shang2020deep, SUN2022102662, Li2023-og} and RCNN-based models for object and instance segmentation~\cite{ocer2020, he2018mask, rs12081288, xiangshu2021, Santos2019-bx, deepforest}. More recently, there has been increased interest in transformer-based models ~\citep{gibril2023large, treeformer, transformers} and general-purpose segmentation models like Segment Anything (SAM)~\cite{kirillov2023segment, OSCO2023103540}. In the last year, several foundational models, trained on large quantities of geospatial data, have been released~\citep{Prithvi-100M-preprint, claymodel, TOLAN2024113888} and show potential for tree detection purposes.

\paragraph{Datasets} Despite continual and frequent advancements in model architectures, the open data landscape for high resolution tree detection is more limited. From the literature listed in the previous section, very few works have released their training data. One effort to collate open datasets from the literature is the MillionTrees project which aims to create a unified benchmark for tree detection across task types~\cite{milliontrees}. Still, the majority of the datasets listed on MillionTrees are limited to, at best, a single country. Interestingly, almost all the datasets currently listed on MillionTrees were published in the last 2-3 years which suggests that attitudes to data release are improving within the community.

A notable open dataset is NeonTreeEvaluation~\cite{ben_weinstein_2022_5914554}, published alongside DeepForest, which is derived from the National Ecological Observation Network (NEON)~\cite{neonnetwork} in the USA. NeonTreeEvaluation includes bounding-box training data from several distinct sites and includes aerial RGB, LIDAR and hyperspectral imaging. This dataset was used to train DeepForest and was released as a benchmark. Other relevant works like DetecTree2 (instance segmentation)~\cite{ball2023} and AutoArborist~\cite{beery2022auto} provide partial training data or require an access request.

As more governmental organisations release open-access geospatial data there is an opportunity to construct benchmark datasets without needing to acquire labels (for example LIDAR surveys, tree inventory, land-cover maps). In this work we combine a variety of data products that are published by the city of Zurich and Switzerland's federal mapping institute to assess our models. This is an approach used by others: DetecTree's~\cite{Bosch2020} semantic segmentation model was trained on an earlier release of this imagery and~\cite{Li2023-og} take advantage of state-acquired imagery over Finland and Denmark. Alongside NEON, other works have used NAIP (National Agricultural Imaging Program) data at 60 cm, also over the US~\cite{naip}. Tree location databases may also be used as keypoint ground truth for density estimation or counting, but are less useful for area/coverage prediction where crown size is required.

To summarise, we find that (1) a large variety of ML models have been employed for tree detection; (2) most works do not share their data and most only evaluate on a small geographic region. We designed our dataset to address these issues, in particular the absence of a global instance-level tree detection dataset.

\section{OAM-TCD Dataset}

\begin{figure}
    \centering
    \includegraphics[width=\linewidth]{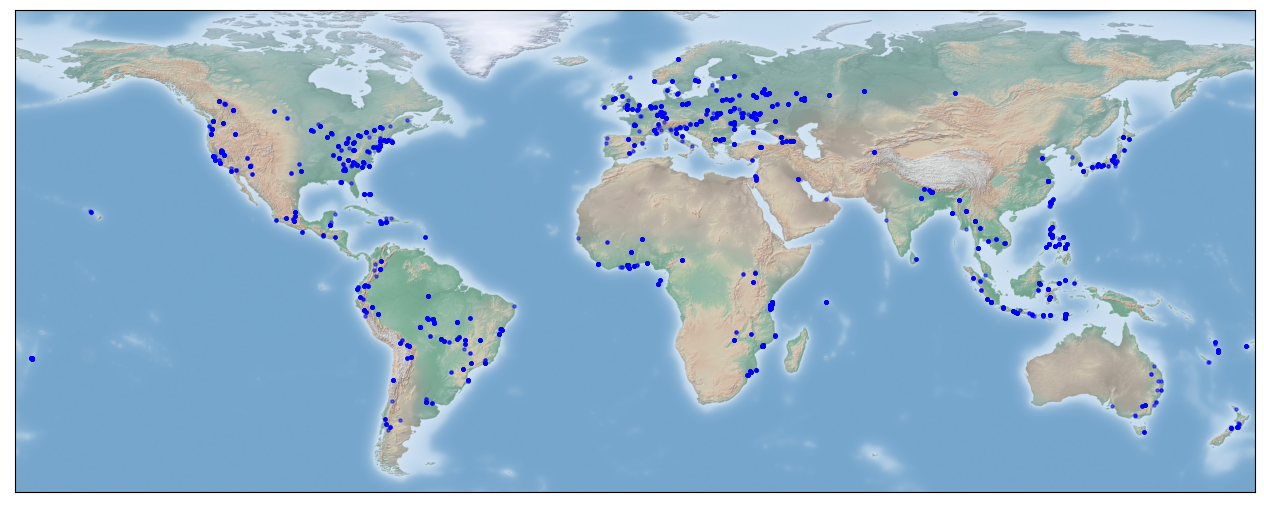}
    \caption{Geospatial distribution of imagery in the dataset. It is clear that some locations are under-represented, but among open-access data, we believe OAM-TCD is the most geographically diverse of its type. The lack of imagery from some regions is due to inherent biases in the data that are uploaded to OAM.}
    \label{fig:annotation_distribution}
\end{figure}

Our dataset - OAM-TCD - is derived from imagery obtained from OpenAerialMap (OAM)~\cite{openaerialmap} and contains instance-level (polygon) annotations for individual and groups of trees. OAM is a repository of permissively licensed global aerial imagery, much of which is user-contributed from UAV-based surveys, alongside some public-domain satellite data (typically from post-disaster events). Sample image tiles from OAM-TCD are shown in Figure~\ref{fig:annotation_examples_main} and a map of the geospatial distribution of the entire dataset is shown in Figure~\ref{fig:annotation_distribution}. Our aim is to provide the community with an open dataset to support tree mapping in global contexts while allowing users to experiment with different mapping tasks. By providing polygon annotations, we provide more flexibility over datasets that only contain keypoints (crown center locations) or bounding boxes.

\subsection{Stratification and diversity}

Our dataset was constructed by repeatedly sampling OAM images from a 1-degree binned world map. Each image was split into tiles which were randomly sampled with some simple filtering to avoid completely empty regions. The number of tiles was constrained by labelling budget and so we report results using a k-fold biome-stratified cross-validation and a test (holdout) split. We partitioned 10\% of the data for holdout testing, stratified such that the biome distribution in the train/test splits are approximately balanced. We then split the remaining 90\% (train data) into 5 folds, also stratified by biome. There are no overlaps between source orthomosaics in the splits to avoid train/test leakage. To reduce file size, we distribute train and test splits as single MS-COCO formatted JSON files and provide the filenames/OAM IDs used for each split; a parallel release on HuggingFace contains the validation fold number in each Parquet entry and we provide a script to generate the training folds.

We also consider biome distribution as a measure of diversity. The WWF Terrestrial Ecoregions of the World~\cite{ecoregions} describe a hierarchical classification for regions on Earth that broadly align with common species distribution. The authors propose 867 ecoregions which are grouped into 14 biomes. In our dataset, we tag each tile with the biome index for the site, if it could be determined. This provides a coarse measure of the types of region which are over- or under-represented in our data. Some biomes, such as \textit{(6) Boreal Forests/Taiga}, \textit{(9) Flooded Grasslands and Savannahs} and \textit{(11) Tundra} are not well-represented in the dataset due to limited availability and spatial biases of imagery in OAM, as well as the sampling technique used to select imagery. More information can be found in the supplementary material.

\subsection{Image and label characteristics}

We provide annotations for tiles of 2048 x 2048 pixels, uniformly resampled to 10 cm/px resolution. Individual tiles are random crops of ``parent" surveys at higher than or equal to the target resolution. Each tile is a geo-referenced image in GeoTIFF format using the EPSG:3395 (World Mercator) projected coordinate reference system (CRS). Imagery is stored in 8-bit 3-channel RGB format with JPEG compression. 10 cm resolution was chosen as a trade-off between object visibility (ease of seeing trees and canopy structure) and ease of labelling with standard annotation tools. The large tile size allows for down-scaling to 0.8 m resolution at a tile size of 256 px, or up-scaling by re-sampling from parent imagery (when resolution permits). Thus, the dataset might also be used for assessing methods for satellite imagery analysis at approximately meter-scale resolution.

Labels are provided as semantic masks and in MS-COCO instance segmentation (polygon) format with two classes: \texttt{tree} and \texttt{canopy} (group of trees). This allows for different modeling approaches to be tested: individual tree and canopy region instance segmentation, or binary tree canopy semantic segmentation. An advantage of polygonal or semantic representations over bounding boxes is that predictions explicitly capture morphological information (i.e. crown shape) and makes area coverage estimation trivial. We did not consider acquisition time when sampling images which has an implication for deciduous trees which exhibit seasonal leaf cover variation. The dataset contains images of trees with and without leaves, and a few images that do not contain any annotations. The compressed dataset volume is 3.9 GB.

A key modeling decision for individual tree detection is whether to attempt to individually segment trees in regions of closed canopy (i.e. where trees are touching). Natural, established, forests have complex structure and it can be extremely difficult to delineate individuals in dense canopy, especially when multiple trees of the same species stand adjacent to each other. With other data like LIDAR or hyperspectral imaging, it is possible to exploit variations in canopy height, or leaf reflectance. Given the challenge of unambiguously segmenting individual trees in closed canopy and using RGB imagery alone, we chose to treat regions as a crowd class so that all regions are labelled, leaving the option open to re-assess annotations in the future to provide a more detailed segmentation.

Approaches in the literature vary. \cite{ben_weinstein_2022_5914554} provide bounding box annotations of trees derived from LIDAR surveys with a smaller set of human annotations. \cite{Brandt2020} segment trees within canopy, train a semantic segmentation model, and use a post-processing step to split detections. ~\cite{ball2023} do not attempt to annotate all trees, but do label individual instances in dense rainforest.~\cite{Li2023-og} train a model with a semantic segmentation and head using a weighted loss function which penalises contact between segmented regions. Some of these works show compelling results from relatively few instance annotations, O(1-10k), which implies it may be possible to selectively fully annotate images in OAM-TCD to achieve similar results.

\subsection{Annotation process}
Our annotation process started with weak labels generated by a Mask-RCNN model trained on a small portion of hand labelled data, followed by human annotation and review. This accelerated the labelling process in cases where the model performed well with little data, for example distinctive species like palms. We hired professional annotators that were not domain experts, but a majority of annotations were reviewed by a second person with ecology expertise. During the labelling process, there was a feedback loop to correct annotations and to answer questions from annotators.

Annotators were given comprehensive guidelines and we revised our process over time in order to address edge cases in labelling. We provide more information on our annotation guidelines in the supplementary material. Annotators were instructed to (1) label all trees in the image that they could confidently identify and (2) if it was not possible to identify individual trees within a group, mark it as ``closed canopy". As a guideline, we suggested that if a region contained fewer than 5 connected trees, they should be annotated individually.

We explored multiple models for annotation pricing, including by overall annotation time, per-polygon and per-image. The sizes of object in our dataset vary from tens of pixels across to close to a full tile size (thousands of pixels) with diverse complexity. Annotation cost varied between 5-10 USD per image depending on the pricing structure and contractor; the cost of labelling the entire dataset was approximately 25k USD.

As our images are not sourced from ecological surveys, in almost all cases we do not have field ground truth available. We therefore aimed for conservative annotation - if it was not obvious whether a tree was an individual or multiple, we asked annotators to mark it as a canopy. However, when considering the labels as purely binary masks, we believe that our labels are generally self-consistent and are high quality.

\subsection{Licensing and access}

A key consideration for our dataset is permissive and open licensing. OAM declares that all imagery in their repository is licensed under Creative Commons Attribution 4.0 International (CC BY 4.0); however a subset of around 10\% of the dataset images are labeled CC BY-NC 4.0 or CC BY-SA 4.0. We re-distribute image tiles under the same license as the metadata (as provided by OAM), which contains attribution information and links to the source orthomosaic. We split the dataset by common image license, so users may choose which combination is most appropriate for their application. In most cases, using only the CC BY imagery should yield good results; we reserve CC BY-SA imagery for testing only. For more information, see the supplementary material.

Our training and detection pipeline is hosted on GitHub (\url{https://github.com/Restor-Foundation/tcd} under an Apache 2.0 license. We provide a hosted version of the dataset on HuggingFace Hub in Apache Parquet format (\url{https://huggingface.co/restor/tcd}), as well as a mirror on Zenodo in MS-COCO format. The dataset is provided as ``ML-ready" and can be used with off-the-shelf object detection frameworks with little to no modification.

\section{Baseline Models}

We trained exemplary models for instance and binary semantic segmentation and release them as usable products for the community. As there are a huge range of model architectures available, we limited training to representative examples of each task type. In order to assess model performance, we report metrics computed on the dataset (k-fold cross validation and holdout split); we also report quantitative and qualitative evaluations on independent data. There is limited like-for-like data available for third party testing available so we constructed tests that demonstrate the strengths and weaknesses of the models. Our release models are trained on all training data, with metrics computed on the holdout set.

Output examples of segmentation results from our models on independent data are shown in Figure~\ref{fig:prediction_examples} and discussed in the following section. A more detailed overview, including model cards, is included in the supplementary information and our GitHub repository.

\paragraph{Dataset pre-processing and augmentation} All models were trained on random crops of 1024x1024 pixels, to provide a large spatial context of around 100 m$^2$ for predictions. We perform a series of random augmentation operations and apply normalisation using ImageNet statistics. The augmentation step includes: horizontal and vertical flipping, rotation, blur and colour adjustments (brightness, hue, saturation). Models are trained at a fixed spatial resolution of 10 cm/px and we do not perform tiled inference when testing on the holdout set. In all cases, we fine-tune models previously trained on MS-COCO or ImageNet which allowed faster convergence compared to models trained from scratch. A fixed seed of 42 was used for each training run for repeatability.

\paragraph{Semantic segmentation} We trained the UNet architecture with ResNet34 and 50 backbones implemented by the Segmentation Models Pytorch library~\cite{segmentation_models_pytorch}. We also trained a series of SegFormer models~\cite{xie2021segformer}, with size variants of mit-b0 through mit-b5. Pytorch Lightning~\cite{pytorchlightning} was used as our training backend and training logs (as Tensorboard event files) are released alongside the final models. Hyperparameter details may be found in the supplementary information.

Table~\ref{tab:segmentation-results} shows model performance from cross-validated models and on the holdout dataset; we report several binary metrics: accuracy, F1 and Jaccard Index (IoU) computed using the TorchMetrics library~\cite{torchmetrics}. In general, SegFormer outperforms UNet, but there is limited gain from using larger model variants. Our hypothesis for the plateau in performance is that we are reaching the limits of the dataset in terms of annotation consistency and diversity, but it is encouraging that such good results can be obtained with relatively lightweight architectures. One advantage of UNet is that it is a well-understood architecture and is permissively licensed; SegFormer is released under a more restrictive NVIDIA research license but we expect similar results could be achieved using Pyramid Vision Transformers (PVT) v2 which shares many of the same architecture choices~\cite{pvt2}.

\begin{table}
    \centering
    \begin{tabular}{@{}cccccccc@{}}
    \toprule
    \multirow{2}{*}{} & \multicolumn{3}{c}{5-fold Cross-Validation} & \multicolumn{3}{c}{Holdout}  \\ \midrule 
                                        & IoU     & Acc     & F1     & IoU     & Acc     & F1     \\
    UNet ResNet34                       &  0.842$\pm$0.004 &  0.882$\pm$0.011 & 0.874$\pm$0.002  &  0.838  &   0.883      &    0.871   \\ 
    UNet ResNet50                       &  0.856$\pm$0.007  &  0.872$\pm$0.006  & 0.885$\pm$0.005  & 0.849  & 0.881   &  0.880     \\ 
    SegFormer mit-b0                    & 0.858$\pm$0.011 & 0.870$\pm$0.021 & 0.887$\pm$0.009 & 0.865 & 0.892 & 0.882  \\ 
    SegFormer mit-b1                    & 0.873$\pm$0.009 & 0.888$\pm$0.015 & 0.900$\pm$0.007 & 0.870 & \textbf{0.897} & 0.891   \\ 
    SegFormer mit-b2                    & 0.878$\pm$0.013 & 0.901$\pm$0.019 & 0.904$\pm$0.010 & 0.871 & 0.889 & 0.898   \\ 
    SegFormer mit-b3                    & 0.879$\pm$0.009 & 0.897$\pm$0.023 & 0.905$\pm$0.007 & 0.875 & 0.884 & 0.875  \\ 
    SegFormer mit-b4                    & 0.880$\pm$0.007 & 0.891$\pm$0.017 & 0.905$\pm$0.007 & 0.875 & 0.891 & 0.901    \\
    SegFormer mit-b5                    & \textbf{0.887$\pm$0.003}&\textbf{0.905$\pm$0.006}&\textbf{0.914$\pm$0.002}& \textbf{0.876} & 0.890 &    \textbf{0.902} \\
    \bottomrule
    \end{tabular}
        \caption{Training results for semantic segmentation models, using the cross-validation and holdout splits in the OAM-TCD dataset. We report the mean and standard deviation of results for cross-validation folds.}
        \label{tab:segmentation-results}
\end{table}

\paragraph{Instance segmentation} For instance segmentation, we trained Mask-RCNN models with a ResNet50 backbone, using the Detectron2 framework~\cite{wu2019detectron2} (\texttt{restor/tcd-mask-rcnn-r50} on HuggingFace). Models were trained with mostly default hyperparameters, with a fixed number of iterations (100,000) at batch size 8, taking the best periodic checkpoint as the final version. We used a tuned learning rate of 0.001 with a stepped schedule following the original paper: a learning rate multiplier of 0.1 applied at iterations 80000 and 90000. Due to high object density in the images, we increased the number of predictions to 512. Cross-validation performance of the models is mAP50={41.79}$\pm${1.38}, and mAP50=43.22 on the holdout set using the COCO API \texttt{segm} task. For comparison, DeepForest report baseline results of mAP50=50 on NeonTreeEvaluation albeit on bounding boxes and not instance predictions; DeepForest also attempts to predict all trees individually.

There are several transformer-based successors to Mask-RCNN, from the DETR~\cite{carion2020endtoend} family of models to Mask2Former~\cite{cheng2022maskedattention}. While we experimented with these architectures, and provide sample training code in the repository, we were unable to reliably outperform Mask-RCNN and found poorer convergence behaviour. Additionally we found there were practical limitations with RAM usage when predicting large numbers of objects, due to implementation/mask representation details. Since tree crowns are - to first order - simple, convex and non-overlapping shapes (deformed circles), there may be value in exploring other memory-efficient architectures that explicitly predict polygon representations, such as DPPD~\cite{zheng2023dppddeformablepolarpolygon} or Poly-YOLO~\cite{hurtik2020polyyolohigherspeedprecise}.

\paragraph{Environmental impact} Our release models were all trained on a single NVIDIA RTX3090 GPU with 24 GB of VRAM at a reduced power target of 200 W. Additional experiments and dataset generation were performed on Google Cloud Platform and ETH's Euler compute cluster. The largest models took approximately 1 day to train. Using the MachineLearning Impact calculator~\cite{lacoste2019quantifying}, we estimate 3.63 kg CO$_2$ equivalent per model variant. This does not take into account testing and failed model runs and so underestimates the overall emission cost of the project, however this can be used as a guideline for emissions generated when training further models.

\section{Third-party benchmarks}

\begin{sidewaysfigure}
    \centering
    \includegraphics[width=\textwidth]{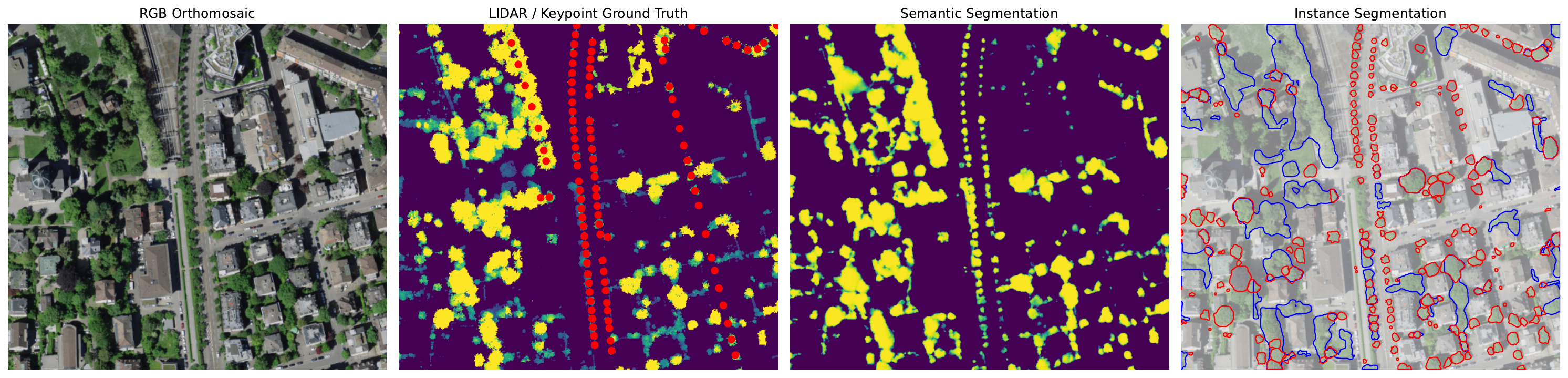}\hfill
    \includegraphics[width=\textwidth]{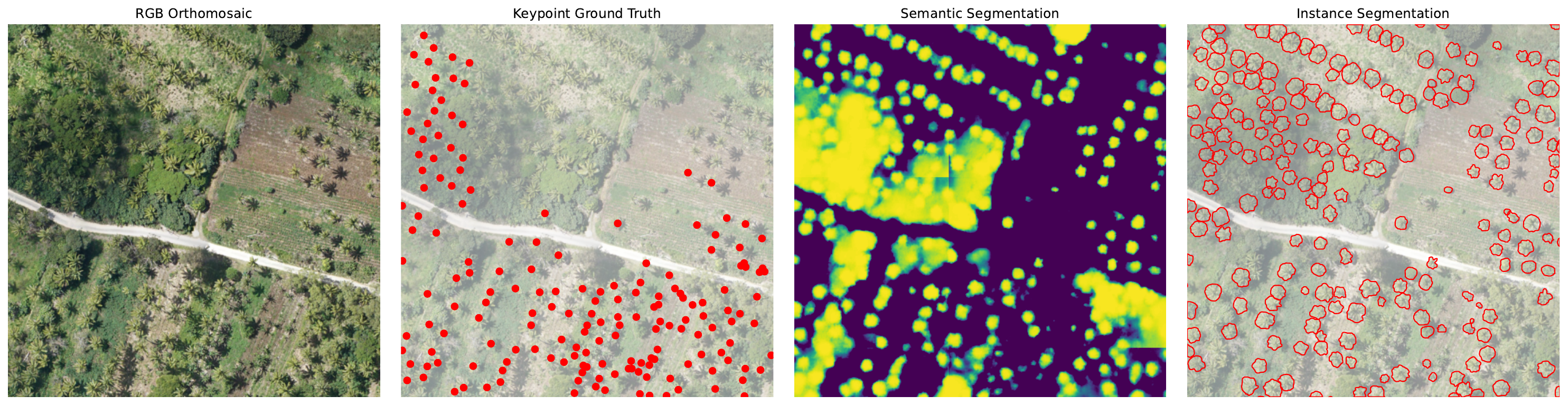}\hfill
    \caption{Model predictions for independent benchmark data. Upper series shows predictions over Zurich in reference to a ground truth canopy height model and tree keypoints, lower series shows predictions on a large orthomosaic from the Kingdom of Tonga with ground truth tree keypoints. Red instances are predicted trees, blue instances are predicted groups. In both datasets only some trees have ground truth labels. Most of the missing detections in the Tonga example above are from small Banana (\textit{Musa}) plants. Models used are \texttt{restor/tcd-segformer-mit-b5} (threshold 0.5) and \texttt{restor/mask-rcnn-r50} (threshold 0.4). Plotting code may be found in the pipeline repository.}
    \label{fig:prediction_examples}
\end{sidewaysfigure}

To assess our models on out-of-domain data, and for practical applications, we report results on two independent datasets. For brevity, we report results from the best models here. Examples of predictions from test data are shown in Figure~\ref{fig:prediction_examples}. Geo-spatial data often far exceeds the VRAM capacity of even a large GPU and images must be processed in a tiled fashion; broadly, we follow the recommendations in~\cite{huang2019tiling} for semantic segmentation and we use heuristics to merge overlapping instances (described in the supplementary material).

\paragraph{Urban tree detection - Zurich} The Federal Institute of Topography in Switzerland (Swisstopo) produces 3-year aerial surveys of the entire country, at 10 cm resolution. By combining 2022 imagery with a up-to-date municipal tree inventory from the city of Zurich, we can assess model performance against an accurate ground truth. A LIDAR-derived CHM within the city boundary, an area of approximately 90 km$^2$, is also available. The inventory data also provides tree species and crown diameter for almost 77k trees within the city. We exclude trees which are marked as planted in 2022, as the inventory and survey dates are not coincident, leaving around 71k keypoints.

We produced prediction maps for canopy coverage and individual trees for the city of Zurich. Our segmentation models show excellent agreement with the LIDAR CHM (for \texttt{tcd-segformer-mit-b5}: IoU=0.791, Accuracy=0.922, F1=0.883) and in some cases provide superior delineation of canopy due to the lower resolution of LIDAR and potentially inaccurate point cloud classification (e.g. returns from some street lamps are present in the CHM). This suggests that our segmentation model could be used in a complementary fashion to better classify LIDAR-derived height rasters, particularly in urban environments. Another example of semantic segmentation output is shown in Figure~\ref{fig:zurich_zoom} and a full image of the city can be found in the supplementary material (Figure~\ref{fig:zurich_semantic} in this document).

Results from instance segmentation also show good agreement with the municipal tree inventory. Since our models detect non-municipal trees, it is not possible to measure the false positive rate. We report a recall of 0.36 for keypoints that were matched to tree labels and 0.73 when also including canopy regions. Filtering predictions using a semantic segmentation map appears to be an effective way of removing high confidence false positives like tree shadow. The relatively low recall for keypoint-in-tree predictions is likely because many trees in the city are close together and are predicted as groups. Our model identifies 193k individual trees within the prediction boundary, the vast majority of which are not actively monitored.

\paragraph{WeRobotics Open AI Challenge} This dataset is a tree detection benchmark. The input is a 325 ha orthomosaic captured over the Kingdom of Tonga~\cite{werobotics_challenge}, hosted on OAM. Tree centers for four species were annotated by humans (13402 total keypoints); not all trees in the image are labelled and our model predicts around twice this many instances over the orthomosaic. Similar to our analysis of Zurich for instance segmentation, we check whether trees in the dataset are captured by either tree or canopy polygons from our model output. We report a recall of 0.64 for keypoints that were matched to tree labels and 0.94 when also including canopy regions. Prediction results are shown in Figure~\ref{fig:prediction_examples} for instance and semantic segmentation output.

\section{Ethics, Limitations and Impact}

We hope that the release of this dataset will bring a positive improvement to the ecological monitoring landscape. There are three main limitations that we highlight here. First, we do not provide extensive validations on in-situ field data with this paper and users should not use these models for decision-making purposes without additional assessments on their own data. Second, there are inconsistencies in the human labelling process and while we have tried to address this by double-reviewing a subset of the imagery, there is noise and bias in the dataset. Some biomes are under-represented, as are some geographic regions. Finally, we did not target annotating every tree in our dataset images. Our instance segmentation models are unlikely to perform well in environments with large regions of dense canopy which users may expect. The model and dataset cards provide more information on these limitations and biases.

\paragraph{PII} We include links to metadata with imagery which may link to personally identifiable information (e.g. names and email addresses for rights holder information) for each image, however this is already in the public domain on OAM, provided by the image uploader. We make the assumption that imagery provided through OAM has the permission of the copyright holder and we provide the metadata for attribution purposes. The images themselves are too low resolution to contain PII (for example human faces).

\section{Conclusion and further work}
There are several avenues for future work: we hope to improve biome coverage in the dataset by including more imagery from under-represented regions; there is ongoing work to improve the consistency of the annotations, for example identifying group labels that could be split into individuals. We are facilitating an ongoing citizen science campaign~\cite{zooniverse} which aims to provide an assessment of inter-annotator agreement and confidence for individual labels. Some of these improvements could also be posed as challenges - for example identifying mislabeled objects.

On its own, we believe that OAM-TCD is the largest open dataset of its kind and will provide other researchers with a novel benchmark for high resolution tree detection, for example zero-shot evaluation of foundational Earth Observation (EO) models. We release our prediction pipeline as a user-friendly tool for tree mapping and hope it will be of use to the ecological community.

\section*{Acknowledgements}
This project was developed through a collaboration between the Restor Foundation and the DS3Lab at ETH Zurich. Funding for research staff and data labelling was supported by a Google.org AI for Social Good grant (ID: TF2012-096892, AI and ML for advancing the monitoring of Forest Restoration).

\bibliography{references}{}

\begin{thebibliography}{10}

\bibitem{green_zurich}
{City of Zurich - Urban Tree Plan (in German)}.
\newblock
  \url{https://www.stadt-zuerich.ch/ted/de/index/gsz/natur-erleben/stadtbaeume/fachplanung-stadtbaeume.html}.
\newblock [Online; acccessed 1-Jun-2024].

\bibitem{globalforestwatch}
{Global Forest Watch. World Resources Institute.}
\newblock \url{http://www.globalforestwatch.org/}, 2014.
\newblock [Online; accessed 1-Jun-2024].

\bibitem{werobotics_challenge}
{Open AI Challenge: Aerial Imagery of South Pacific Islands}.
\newblock \url{https://werobotics.org/blog/open-ai-challenge-2}, 2018.
\newblock [Online; accessed 1-Jun-2024].

\bibitem{claymodel}
Clay foundation model.
\newblock \url{https://clay-foundation.github.io/model/}, 2023.
\newblock [Online; accessed 1-Jun-2024].

\bibitem{milliontrees}
{MillionTrees: A Benchmark Dataset for Airborne Machine Learning}.
\newblock \url{https://milliontrees.idtrees.org/}, 2023.
\newblock [Online; accessed 1-Jun-2024].

\bibitem{neonnetwork}
{NEON (National Ecological Observatory Network)}.
\newblock \url{https://www.neonscience.org/}, 2024.
\newblock [Online; accessed 1-Jun-2024].

\bibitem{norwaynicfi}
{Norway’s International Climate and Forest Initiative (NICFI)}.
\newblock \url{https://www.nicfi.no/}, 2024.
\newblock [Online; accessed 1-Jun-2024].

\bibitem{openaerialmap}
Openaerialmap.
\newblock \url{https://openaerialmap.org/}, 2024.
\newblock [Online; acccessed 1-Jun-2024].

\bibitem{zooniverse}
{Tag Trees, Zooniverse}.
\newblock \url{https://www.zooniverse.org/projects/physics-josh/tag-trees},
  2024.
\newblock [Ongoing campaign, online; acccessed 1-Jun-2024].

\bibitem{treeformer}
Hamed~Amini Amirkolaee, Miaojing Shi, and Mark Mulligan.
\newblock Treeformer: A semi-supervised transformer-based framework for tree
  counting from a single high-resolution image.
\newblock {\em IEEE Transactions on Geoscience and Remote Sensing}, 61:1--15,
  2023.

\bibitem{ball2023}
James G.~C. Ball, Sebastian H.~M. Hickman, Tobias~D. Jackson, Xian~Jing Koay,
  James Hirst, et~al.
\newblock Accurate delineation of individual tree crowns in tropical forests
  from aerial rgb imagery using mask r-cnn.
\newblock {\em Remote Sensing in Ecology and Conservation}, 9(5):641--655,
  2023.

\bibitem{crowther2019}
Jean-Francois Bastin, Yelena Finegold, Claude Garcia, Danilo Mollicone, Marcelo
  Rezende, et~al.
\newblock {The global tree restoration potential}.
\newblock {\em Science}, 365(6448):76--79, 2019.

\bibitem{beery2022auto}
Sara Beery, Guanhang Wu, Trevor Edwards, Filip Pavetic, Bo~Majewski, et~al.
\newblock The auto arborist dataset: A large-scale benchmark for multiview
  urban forest monitoring under domain shift.
\newblock In {\em Proceedings of the IEEE/CVF Conference on Computer Vision and
  Pattern Recognition}, pages 21294--21307, 2022.

\bibitem{bendavid2010}
Shai Ben-David, John Blitzer, Koby Crammer, Alex Kulesza, Fernando Pereira, and
  Jennifer~Wortman Vaughan.
\newblock {A theory of learning from different domains}.
\newblock {\em Machine Learning}, 79(1-2):151--175, 2010.

\bibitem{benhammou2022sentinel2globallulc}
Yassir Benhammou, Domingo Alcaraz-Segura, Emilio Guirado, Rohaifa Khaldi,
  et~al.
\newblock Sentinel2globallulc: A sentinel-2 rgb image tile dataset for global
  land use/cover mapping with deep learning.
\newblock {\em Scientific Data}, 9(1):681, 2022.

\bibitem{Bosch2020}
Martí Bosch.
\newblock Detectree: Tree detection from aerial imagery in python.
\newblock {\em Journal of Open Source Software}, 5(50):2172, 2020.

\bibitem{BRANDT2023113574}
John Brandt, Jessica Ertel, Justine Spore, and Fred Stolle.
\newblock Wall-to-wall mapping of tree extent in the tropics with sentinel-1
  and sentinel-2.
\newblock {\em Remote Sensing of Environment}, 292:113574, 2023.

\bibitem{Brandt2020}
Martin Brandt, Compton~J. Tucker, Ankit Kariryaa, Kjeld Rasmussen, Christin
  Abel, et~al.
\newblock An unexpectedly large count of trees in the west african sahara and
  sahel.
\newblock {\em Nature}, 587(7832):78--82, Nov 2020.

\bibitem{cao2019comparison}
Lin Cao, Hao Liu, Xiaoyao Fu, Zhengnan Zhang, Xin Shen, and Honghua Ruan.
\newblock Comparison of uav lidar and digital aerial photogrammetry point
  clouds for estimating forest structural attributes in subtropical planted
  forests.
\newblock {\em Forests}, 10(2):145, 2019.

\bibitem{carion2020endtoend}
Nicolas Carion, Francisco Massa, Gabriel Synnaeve, Nicolas Usunier, Alexander
  Kirillov, and Sergey Zagoruyko.
\newblock End-to-end object detection with transformers, 2020.

\bibitem{carrio2017review}
Adrian Carrio, Carlos Sampedro, Alejandro Rodriguez-Ramos, and Pascual Campoy.
\newblock A review of deep learning methods and applications for unmanned
  aerial vehicles.
\newblock {\em Journal of Sensors}, 2017, 2017.

\bibitem{cazzolla2022number}
Roberto Cazzolla~Gatti, Peter~B Reich, Javier~GP Gamarra, Tom Crowther, Cang
  Hui, et~al.
\newblock The number of tree species on earth.
\newblock {\em Proceedings of the National Academy of Sciences},
  119(6):e2115329119, 2022.

\bibitem{mmdetection}
Kai Chen, Jiaqi Wang, Jiangmiao Pang, Yuhang Cao, Yu~Xiong, et~al.
\newblock {MMDetection}: Open mmlab detection toolbox and benchmark.
\newblock {\em arXiv preprint arXiv:1906.07155}, 2019.

\bibitem{cheng2022maskedattention}
Bowen Cheng, Ishan Misra, Alexander~G. Schwing, Alexander Kirillov, and Rohit
  Girdhar.
\newblock Masked-attention mask transformer for universal image segmentation.
\newblock {\em arXiv preprint arXiv:2112.01527}, 2022.

\bibitem{CROWTHER2022476}
Thomas~W. Crowther, Stephen~M. Thomas, Johan {van den Hoogen}, Niamh Robmann,
  Alfredo Chavarría, Andrew Cottam, et~al.
\newblock Restor: Transparency and connectivity for the global environmental
  movement.
\newblock {\em One Earth}, 5(5):476--481, 2022.

\bibitem{drones2040039}
Ovidiu Csillik, John Cherbini, Robert Johnson, Andy Lyons, and Maggi Kelly.
\newblock Identification of citrus trees from unmanned aerial vehicle imagery
  using convolutional neural networks.
\newblock {\em Drones}, 2(4), 2018.

\bibitem{torchmetrics}
Nicki~Skafte Detlefsen, Jiri Borovec, Justus Schock, Ananya~Harsh Jha, Teddy
  Koker, et~al.
\newblock Torchmetrics - measuring reproducibility in pytorch.
\newblock {\em Journal of Open Source Software}, 7(70):4101, 2022.

\bibitem{runmin2020}
Runmin Dong, Weijia Li, Haohuan Fu, Lin Gan, Le~Yu, et~al.
\newblock {Oil palm plantation mapping from high-resolution remote sensing
  images using deep learning}.
\newblock {\em International Journal of Remote Sensing}, 41(5):2022--2046,
  2020.

\bibitem{drusch2012sentinel}
Matthias Drusch, Umberto Del~Bello, S{\'e}bastien Carlier, Olivier Colin,
  Veronica Fernandez, Ferran Gascon, Bianca Hoersch, Claudia Isola, Paolo
  Laberinti, Philippe Martimort, et~al.
\newblock Sentinel-2: Esa's optical high-resolution mission for gmes
  operational services.
\newblock {\em Remote sensing of Environment}, 120:25--36, 2012.

\bibitem{Dubayah_2020}
Ralph Dubayah, James~Bryan Blair, Scott Goetz, Lola Fatoyinbo, Matthew Hansen,
  et~al.
\newblock The global ecosystem dynamics investigation: High-resolution laser
  ranging of the earth’s forests and topography.
\newblock {\em Science of Remote Sensing}, 1:100002, June 2020.

\bibitem{dwyer1991significance}
John~F Dwyer, Herbert~W Schroeder, and Paul~H Gobster.
\newblock The significance of urban trees and forests: toward a deeper
  understanding of values.
\newblock {\em Journal of Arboriculture}, 17(10):276--284, 1991.

\bibitem{fagan2020lesson}
Matthew~E Fagan.
\newblock A lesson unlearned? underestimating tree cover in drylands biases
  global restoration maps.
\newblock {\em Global Change Biology}, 26(9):4679--4690, 2020.

\bibitem{pytorchlightning}
William Falcon and The PyTorch~Lightning team.
\newblock Pytorch lightning.
\newblock \url{https://doi.org/10.5281/zenodo.11245620}, May 2024.

\bibitem{freudenberg2019}
Maximilian Freudenberg, Nils Nölke, Alejandro Agostini, Kira Urban, Florentin
  Wörgötter, and Christoph Kleinn.
\newblock {Large Scale Palm Tree Detection In High Resolution Satellite Images
  Using U-Net}.
\newblock {\em Remote Sensing}, 11(3):312, 2019.

\bibitem{rs12081288}
José~R. G.~Braga, Vinícius Peripato, Ricardo Dalagnol, Matheus P.~Ferreira,
  Yuliya Tarabalka, et~al.
\newblock Tree crown delineation algorithm based on a convolutional neural
  network.
\newblock {\em Remote Sensing}, 12(8), 2020.

\bibitem{ganz2019measuring}
Selina Ganz, Yannek K{\"a}ber, and Petra Adler.
\newblock Measuring tree height with remote sensing—a comparison of
  photogrammetric and lidar data with different field measurements.
\newblock {\em Forests}, 10(8):694, 2019.

\bibitem{gdal}
{GDAL/OGR contributors}.
\newblock {GDAL/OGR} geospatial data abstraction software library.
\newblock \url{https://gdal.org}, 2024.

\bibitem{geos}
{GEOS contributors}.
\newblock {\em {GEOS} computational geometry library}.
\newblock Open Source Geospatial Foundation, 2021.

\bibitem{gibril2023large}
Mohamed Barakat~A Gibril, Helmi Zulhaidi~Mohd Shafri, Rami Al-Ruzouq, Abdallah
  Shanableh, Faten Nahas, and Saeed Al~Mansoori.
\newblock Large-scale date palm tree segmentation from multiscale uav-based and
  aerial images using deep vision transformers.
\newblock {\em Drones}, 7(2):93, 2023.

\bibitem{fiona}
Sean Gillies, René Buffat, Joshua Arnott, Mike~W. Taves, Kevin Wurster, et~al.
\newblock Fiona.
\newblock \url{https://github.com/Toblerity/Fiona}, 10 2023.

\bibitem{rtree}
Sean Gillies, Howard Butler, Brent Pederson, Adam Steward, Taves, et~al.
\newblock libspatialindex/rtree.
\newblock \url{https://github.com/Toblerity/rtree}, 2024.

\bibitem{gillies_2019}
Sean Gillies et~al.
\newblock Rasterio: geospatial raster i/o for {Python} programmers.
\newblock \url{https://github.com/rasterio/rasterio}, 2013--.

\bibitem{shapely}
Sean Gillies, Casper van~der Wel, Joris Van~den Bossche, Mike~W. Taves, Joshua
  Arnott, Brendan~C. Ward, et~al.
\newblock Shapely.
\newblock \url{https://github.com/shapely/shapely}, 4 2024.

\bibitem{rs9121220}
Emilio Guirado, Siham Tabik, Domingo Alcaraz-Segura, Javier Cabello, and
  Francisco Herrera.
\newblock Deep-learning versus obia for scattered shrub detection with google
  earth imagery: Ziziphus lotus as case study.
\newblock {\em Remote Sensing}, 9(12), 2017.

\bibitem{hansen2013high}
Matthew~C Hansen, Peter~V Potapov, Rebecca Moore, Matt Hancher, Svetlana~A
  Turubanova, et~al.
\newblock High-resolution global maps of 21st-century forest cover change.
\newblock {\em Science}, 342(6160):850--853, 2013.

\bibitem{he2018mask}
Kaiming He, Georgia Gkioxari, Piotr Doll{\'a}r, and Ross Girshick.
\newblock {Mask R-CNN}.
\newblock In {\em Proceedings of the IEEE international conference on computer
  vision}, pages 2961--2969, 2017.

\bibitem{huang2019tiling}
Bohao Huang, Daniel Reichman, Leslie~M. Collins, Kyle Bradbury, and Jordan~M.
  Malof.
\newblock Tiling and stitching segmentation output for remote sensing: Basic
  challenges and recommendations.
\newblock {\em arXiv preprint arXiv:1805.12219}, 2019.

\bibitem{hurtik2020polyyolohigherspeedprecise}
Petr Hurtik, Vojtech Molek, Jan Hula, Marek Vajgl, Pavel Vlasanek, and Tomas
  Nejezchleba.
\newblock {Poly-YOLO: higher speed, more precise detection and instance
  segmentation for YOLOv3}, 2020.

\bibitem{segmentation_models_pytorch}
Pavel Iakubovskii.
\newblock Segmentation models pytorch.
\newblock \url{https://github.com/qubvel/segmentation_models.pytorch}, 2019.

\bibitem{Prithvi-100M-preprint}
Johannes Jakubik, Sujit Roy, C.~E. Phillips, Paolo Fraccaro, Denys Godwin,
  et~al.
\newblock {Foundation Models for Generalist Geospatial Artificial
  Intelligence}.
\newblock {\em Preprint Available on arxiv:2310.18660}, October 2023.

\bibitem{2011.Ke.International}
Yinghai Ke and Lindi~J. Quackenbush.
\newblock A review of methods for automatic individual tree-crown detection and
  delineation from passive remote sensing.
\newblock {\em International Journal of Remote Sensing}, 32(17):4725--4747,
  2011.

\bibitem{kirillov2023segment}
Alexander Kirillov, Eric Mintun, Nikhila Ravi, Hanzi Mao, Chloe Rolland, et~al.
\newblock Segment anything.
\newblock In {\em Proceedings of the IEEE/CVF International Conference on
  Computer Vision}, pages 4015--4026, 2023.

\bibitem{pmlr-v139-koh21a}
Pang~Wei Koh, Shiori Sagawa, Henrik Marklund, Sang~Michael Xie, Marvin Zhang,
  et~al.
\newblock Wilds: A benchmark of in-the-wild distribution shifts.
\newblock In Marina Meila and Tong Zhang, editors, {\em Proceedings of the 38th
  International Conference on Machine Learning}, volume 139 of {\em Proceedings
  of Machine Learning Research}, pages 5637--5664. PMLR, 18--24 Jul 2021.

\bibitem{NIPS2012_c399862d}
Alex Krizhevsky, Ilya Sutskever, and Geoffrey~E Hinton.
\newblock Imagenet classification with deep convolutional neural networks.
\newblock In F.~Pereira, C.J. Burges, L.~Bottou, and K.Q. Weinberger, editors,
  {\em Advances in Neural Information Processing Systems}, volume~25. Curran
  Associates, Inc., 2012.

\bibitem{lacoste2024geo}
Alexandre Lacoste, Nils Lehmann, Pau Rodriguez, Evan Sherwin, Hannah Kerner,
  Bj{\"o}rn L{\"u}tjens, Jeremy Irvin, David Dao, Hamed Alemohammad, Alexandre
  Drouin, et~al.
\newblock Geo-bench: Toward foundation models for earth monitoring.
\newblock {\em Advances in Neural Information Processing Systems}, 36, 2024.

\bibitem{lacoste2019quantifying}
Alexandre Lacoste, Alexandra Luccioni, Victor Schmidt, and Thomas Dandres.
\newblock Quantifying the carbon emissions of machine learning.
\newblock {\em arXiv preprint arXiv:1910.09700}, 2019.

\bibitem{lang2023high}
Nico Lang, Walter Jetz, Konrad Schindler, and Jan~Dirk Wegner.
\newblock A high-resolution canopy height model of the earth.
\newblock {\em Nature Ecology \& Evolution}, pages 1--12, 2023.

\bibitem{lecunn2015}
Yann LeCun, Yoshua Bengio, and Geoffrey Hinton.
\newblock {Deep learning}.
\newblock {\em Nature}, 521(7553):436--444, 2015.

\bibitem{lewis2019restoring}
Simon~L Lewis, Charlotte~E Wheeler, Edward~TA Mitchard, and Alexander Koch.
\newblock Restoring natural forests is the best way to remove atmospheric
  carbon.
\newblock {\em Nature}, 568(7750):25--28, 2019.

\bibitem{Li2023-og}
Sizhuo Li, Martin Brandt, Rasmus Fensholt, Ankit Kariryaa, Christian Igel,
  et~al.
\newblock Deep learning enables image-based tree counting, crown segmentation,
  and height prediction at national scale.
\newblock {\em PNAS Nexus}, 2(4):gad076, April 2023.

\bibitem{rs9010022}
Weijia Li, Haohuan Fu, Le~Yu, and Arthur Cracknell.
\newblock Deep learning based oil palm tree detection and counting for
  high-resolution remote sensing images.
\newblock {\em Remote Sensing}, 9(1), 2017.

\bibitem{LINDENMAYER20101317}
David~B. Lindenmayer and Gene~E. Likens.
\newblock The science and application of ecological monitoring.
\newblock {\em Biological Conservation}, 143(6):1317--1328, 2010.

\bibitem{Lindenmayer2022-tn}
David~B Lindenmayer, John Woinarski, Sarah Legge, Martine Maron, Stephen~T
  Garnett, et~al.
\newblock Eight things you should never do in a monitoring program: an
  australian perspective.
\newblock {\em Environ. Monit. Assess.}, 194(10):701, August 2022.

\bibitem{malek2014efficient}
Salim Malek, Yakoub Bazi, Naif Alajlan, Haikel AlHichri, and Farid Melgani.
\newblock Efficient framework for palm tree detection in uav images.
\newblock {\em IEEE Journal of Selected Topics in Applied Earth Observations
  and Remote Sensing}, 7(12):4692--4703, 2014.

\bibitem{miraki2021}
Mojdeh Miraki, Hormoz Sohrabi, Parviz Fatehi, and Mathias Kneubuehler.
\newblock {Individual tree crown delineation from high-resolution UAV images in
  broadleaf forest}.
\newblock {\em Ecological Informatics}, 61:101207, 2021.

\bibitem{Mo2023}
Lidong Mo, Constantin~M. Zohner, Peter~B. Reich, Jingjing Liang, Sergio
  de~Miguel, et~al.
\newblock Integrated global assessment of the natural forest carbon potential.
\newblock {\em Nature}, 624(7990):92--101, Dec 2023.

\bibitem{planetnicfi}
Lampros Mouselimis.
\newblock {\em {{PlanetNICFI}: Processing of the 'Planet NICFI' Satellite
  Imagery using R}}, 2022.
\newblock [Online; accessed 1-Jun-2024].

\bibitem{Mugabowindekwe2023-qh}
Maurice Mugabowindekwe, Martin Brandt, J{\'e}r{\^o}me Chave, Florian Reiner,
  David~L Skole, et~al.
\newblock Nation-wide mapping of tree-level aboveground carbon stocks in
  rwanda.
\newblock {\em Nat. Clim. Chang.}, 13(1):91--97, 2023.

\bibitem{nowak1996}
David~J. Nowak, Rowan~A. Rowntree, E.Gregory McPherson, Susan~M. Sisinni,
  Esther~R. Kerkmann, and Jack~C. Stevens.
\newblock {Measuring and analyzing urban tree cover}.
\newblock {\em Landscape and Urban Planning}, 36(1):49--57, 1996.

\bibitem{ocer2020}
Nuri~Erkin Ocer, Gordana Kaplan, Firat Erdem, Dilek~Kucuk Matci, and Ugur
  Avdan.
\newblock Tree extraction from multi-scale uav images using mask r-cnn with
  fpn.
\newblock {\em Remote Sensing Letters}, 11(9):847--856, 2020.

\bibitem{ecoregions}
David~M. Olson, Eric Dinerstein, Eric~D. Wikramanayake, Neil~D. Burgess, et~al.
\newblock {Terrestrial Ecoregions of the World: A New Map of Life on Earth: A
  new global map of terrestrial ecoregions provides an innovative tool for
  conserving biodiversity}.
\newblock {\em BioScience}, 51(11):933--938, 11 2001.

\bibitem{OSCO2023103540}
Lucas~Prado Osco, Qiusheng Wu, Eduardo~Lopes {de Lemos}, Wesley~Nunes
  Gonçalves, Ana Paula~Marques Ramos, et~al.
\newblock The segment anything model (sam) for remote sensing applications:
  From zero to one shot.
\newblock {\em International Journal of Applied Earth Observation and
  Geoinformation}, 124:103540, 2023.

\bibitem{pearse2018comparison}
Grant~D Pearse, Jonathan~P Dash, Henrik~J Persson, and Michael~S Watt.
\newblock Comparison of high-density lidar and satellite photogrammetry for
  forest inventory.
\newblock {\em ISPRS journal of photogrammetry and remote sensing},
  142:257--267, 2018.

\bibitem{scikit-learn}
F.~Pedregosa, G.~Varoquaux, A.~Gramfort, V.~Michel, B.~Thirion, O.~Grisel,
  et~al.
\newblock Scikit-learn: Machine learning in {P}ython.
\newblock {\em Journal of Machine Learning Research}, 12:2825--2830, 2011.

\bibitem{phiri2020}
Darius Phiri, Matamyo Simwanda, Serajis Salekin, Vincent~R. Nyirenda, Yuji
  Murayama, and Manjula Ranagalage.
\newblock {Sentinel-2 Data for Land Cover/Use Mapping: A Review}.
\newblock {\em Remote Sensing}, 12(14):2291, 2020.

\bibitem{rahman2020}
Mohammad~A. Rahman, Laura~M.F. Stratopoulos, Astrid Moser-Reischl, Teresa
  Zölch, Karl-Heinz Häberle, et~al.
\newblock {Traits of trees for cooling urban heat islands: A meta-analysis}.
\newblock {\em Building and Environment}, 170:106606, 2020.

\bibitem{reierson2021}
Gyri Reiersen, David Dao, Bj{\"{o}}rn L{\"{u}}tjens, Konstantin Klemmer,
  Xiaoxiang Zhu, and Ce~Zhang.
\newblock Tackling the overestimation of forest carbon with deep learning and
  aerial imagery.
\newblock {\em CoRR}, abs/2107.11320, 2021.

\bibitem{Reiner2023-wq}
Florian Reiner, Martin Brandt, Xiaoye Tong, David Skole, Ankit Kariryaa, et~al.
\newblock More than one quarter of africa's tree cover is found outside areas
  previously classified as forest.
\newblock {\em Nat. Commun.}, 14(1):2258, May 2023.

\bibitem{ronneberger2015u}
Olaf Ronneberger, Philipp Fischer, and Thomas Brox.
\newblock U-net: Convolutional networks for biomedical image segmentation.
\newblock In {\em Medical Image Computing and Computer-Assisted
  Intervention--MICCAI 2015: 18th International Conference, Munich, Germany,
  October 5-9, 2015, Proceedings, Part III 18}, pages 234--241. Springer, 2015.

\bibitem{safonova2021individual}
Anastasiia Safonova, Yousif Hamad, Egor Dmitriev, Georgi Georgiev, Vladislav
  Trenkin, et~al.
\newblock Individual tree crown delineation for the species classification and
  assessment of vital status of forest stands from uav images.
\newblock {\em Drones}, 5(3):77, 2021.

\bibitem{Santos2019-bx}
Anderson Aparecido~Dos Santos, Jos{\'e} Marcato~Junior, M{\'a}rcio~Santos
  Ara{\'u}jo, David~Robledo Di~Martini, et~al.
\newblock Assessment of {CNN-based} methods for individual tree detection on
  images captured by {RGB} cameras attached to {UAVs}.
\newblock {\em Sensors (Basel)}, 19(16):3595, August 2019.

\bibitem{naip}
National Agriculture Imagery Program (NAIP)~[Data set].
\newblock Earth resources observation and science {(EROS)} center.
\newblock \url{https://doi.org/10.5066/F7QN651G} [Online, accessed 1-Jun-2024],
  2017.

\bibitem{shang2020deep}
Gaogao Shang, Gang Liu, Peng Zhu, Jiangyi Han, Changgao Xia, and Kun Jiang.
\newblock A deep residual u-type network for semantic segmentation of orchard
  environments.
\newblock {\em Applied Sciences}, 11(1):322, 2020.

\bibitem{SUN2022102662}
Ying Sun, Ziming Li, Huagui He, Liang Guo, Xinchang Zhang, and Qinchuan Xin.
\newblock Counting trees in a subtropical mega city using the instance
  segmentation method.
\newblock {\em International Journal of Applied Earth Observation and
  Geoinformation}, 106:102662, 2022.

\bibitem{ter2013hyperdominance}
Hans Ter~Steege, Nigel~CA Pitman, Daniel Sabatier, Christopher Baraloto,
  Rafael~P Salom{\~a}o, et~al.
\newblock Hyperdominance in the amazonian tree flora.
\newblock {\em Science}, 342(6156):1243092, 2013.

\bibitem{TOLAN2024113888}
Jamie Tolan, Hung-I Yang, Benjamin Nosarzewski, Guillaume Couairon, Huy~V. Vo,
  et~al.
\newblock Very high resolution canopy height maps from rgb imagery using
  self-supervised vision transformer and convolutional decoder trained on
  aerial lidar.
\newblock {\em Remote Sensing of Environment}, 300:113888, 2024.

\bibitem{transformers}
Ashish Vaswani, Noam Shazeer, Niki Parmar, Jakob Uszkoreit, Llion Jones, et~al.
\newblock Attention is all you need.
\newblock In I.~Guyon, U.~Von Luxburg, S.~Bengio, H.~Wallach, R.~Fergus,
  S.~Vishwanathan, and R.~Garnett, editors, {\em Advances in Neural Information
  Processing Systems}, volume~30. Curran Associates, Inc., 2017.

\bibitem{pvt2}
Wenhai Wang, Enze Xie, Xiang Li, Deng-Ping Fan, Kaitao Song, Ding Liang, Tong
  Lu, Ping Luo, and Ling Shao.
\newblock Pvt v2: Improved baselines with pyramid vision transformer.
\newblock {\em Computational Visual Media}, 8(3):415–424, March 2022.

\bibitem{wang2020generalizing}
Yaqing Wang, Quanming Yao, James~T Kwok, and Lionel~M Ni.
\newblock Generalizing from a few examples: A survey on few-shot learning.
\newblock {\em ACM computing surveys (csur)}, 53(3):1--34, 2020.

\bibitem{ben_weinstein_2022_5914554}
Ben Weinstein, Sergio Marconi, and Ethan White.
\newblock {Data for the NeonTreeEvaluation Benchmark}.
\newblock \url{https://doi.org/10.5281/zenodo.5914554} [Online, accessed
  1-Jun-2024], January 2022.

\bibitem{deepforest}
Ben~G. Weinstein, Sergio Marconi, Stephanie Bohlman, Alina Zare, and Ethan
  White.
\newblock Individual tree-crown detection in rgb imagery using semi-supervised
  deep learning neural networks.
\newblock {\em Remote Sensing}, 11(11), 2019.

\bibitem{weinstein2020cross}
Ben~G Weinstein, Sergio Marconi, Stephanie~A Bohlman, Alina Zare, and Ethan~P
  White.
\newblock Cross-site learning in deep learning rgb tree crown detection.
\newblock {\em Ecological Informatics}, 56:101061, 2020.

\bibitem{wolf2020urban}
Kathleen~L Wolf, Sharon~T Lam, Jennifer~K McKeen, Gregory~RA Richardson,
  Matilda van~den Bosch, and Adrina~C Bardekjian.
\newblock Urban trees and human health: A scoping review.
\newblock {\em International journal of environmental research and public
  health}, 17(12):4371, 2020.

\bibitem{wu2019detectron2}
Yuxin Wu, Alexander Kirillov, Francisco Massa, Wan-Yen Lo, and Ross Girshick.
\newblock Detectron2.
\newblock \url{https://github.com/facebookresearch/detectron2}, 2019.

\bibitem{wulder2019current}
Michael~A Wulder, Thomas~R Loveland, David~P Roy, Christopher~J Crawford,
  Jeffrey~G Masek, et~al.
\newblock Current status of landsat program, science, and applications.
\newblock {\em Remote sensing of environment}, 225:127--147, 2019.

\bibitem{xiangshu2021}
Xiangshu Xi, Kai Xia, Yinhui Yang, Xiaochen Du, and Hailin Feng.
\newblock {Evaluation of dimensionality reduction methods for individual tree
  crown delineation using instance segmentation network and UAV multispectral
  imagery in urban forest}.
\newblock {\em Computers and Electronics in Agriculture}, 191:106506, 2021.

\bibitem{xie2021segformer}
Enze Xie, Wenhai Wang, Zhiding Yu, Anima Anandkumar, Jose~M Alvarez, and Ping
  Luo.
\newblock Segformer: Simple and efficient design for semantic segmentation with
  transformers.
\newblock In {\em Neural Information Processing Systems (NeurIPS)}, 2021.

\bibitem{rs8040333}
Zhen Zhen, Lindi~J. Quackenbush, and Lianjun Zhang.
\newblock Trends in automatic individual tree crown detection and
  delineation—evolution of lidar data.
\newblock {\em Remote Sensing}, 8(4), 2016.

\bibitem{zheng2023dppddeformablepolarpolygon}
Yang Zheng, Oles Andrienko, Yonglei Zhao, Minwoo Park, and Trung Pham.
\newblock {DPPD: Deformable Polar Polygon Object Detection}, 2023.

\end{thebibliography}

\newpage

\appendix

\section*{Appendix / Supplementary Material}

\begin{figure}[h]
	\centering
	\includegraphics[width=1.0\linewidth]{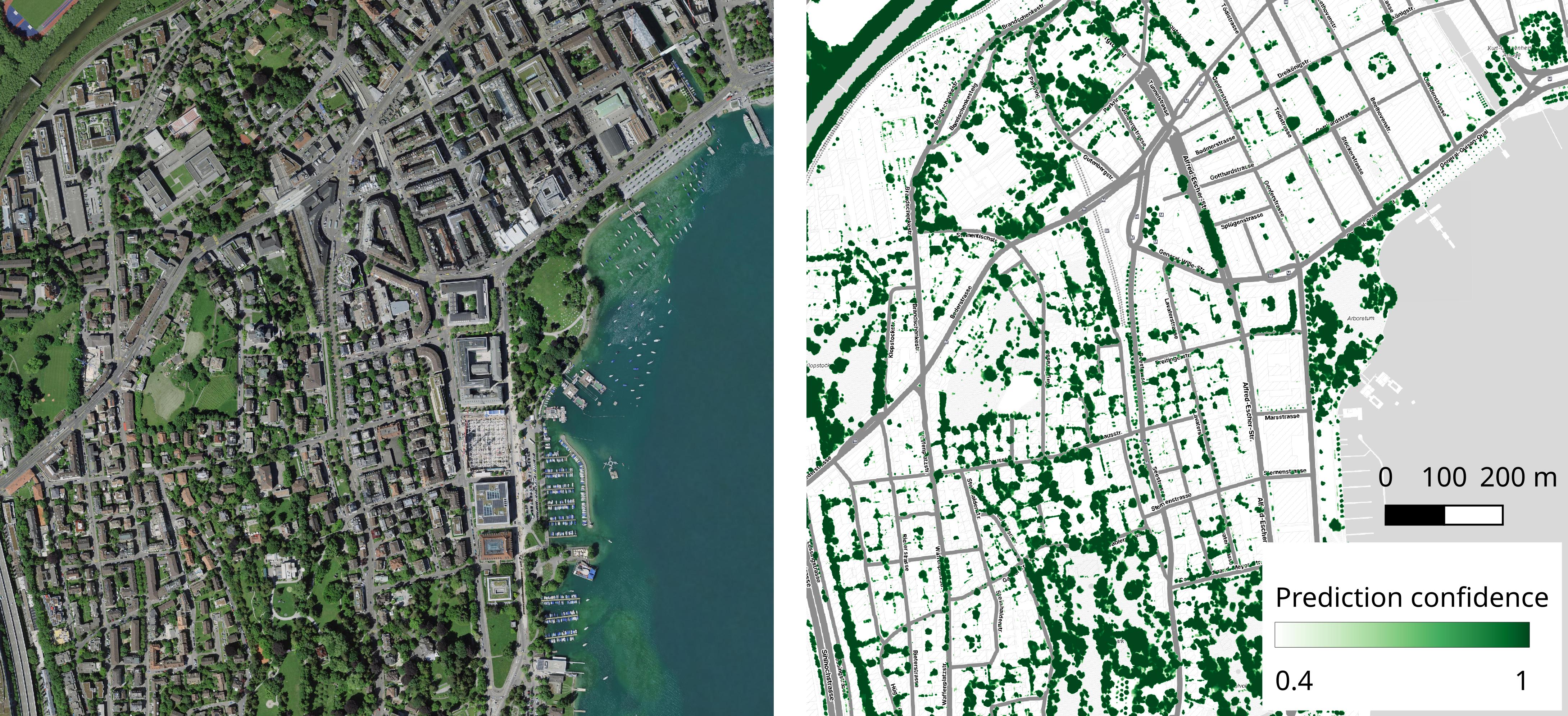}
	\caption{Tree semantic segmentation for Zurich, predicted at 10 cm/px. Predictions with a confidence of < 0.4 are hidden. Left: 10 cm RGB orthomosaic provided by the Swiss Federal Office of Topography swisstopo/SWISSIMAGE 10 cm (2022), Right: prediction heat map. Zooming in is recommended to see small details, e.g. trees along the top edge of the lake.  Base map tiles by Stamen Design, under CC BY 4.0. Data by OpenStreetMap, under ODbL.}
	\label{fig:zurich_zoom}
\end{figure}

\section{Dataset Information}
\label{dataset}
\subsection{Dataset Card}

This dataset card is reproduced from our HuggingFace repository (\url{https://huggingface.co/datasets/restor/tcd}), using the provided template as reference for headings. Minor changes have been made for readability and formatting.

\paragraph{Dataset Details} OAM-TCD is a dataset of high-resolution (10 cm/px) tree cover maps with instance-level masks for 280k trees and 56k tree groups. Images in the dataset are provided as {2048x2048 px} RGB GeoTIFF tiles. The dataset can be used to train both instance segmentation models and semantic segmentation models.

\paragraph{Dataset Description}
\begin{itemize}
	\item Curated by: Restor and ETH Zurich
	\item Funded by: Restor and ETH Zurich, supported by a Google.org AI for Social Good grant (ID: TF2012-096892, AI and ML for advancing the monitoring of Forest Restoration)
	\item License: Annotations are predominantly released under a CC BY 4.0 license, with around 10\% licensed as CC BY-NC 4.0 or CC BY-SA 4.0. These less permissive images are distributed in separate repositories to avoid any ambiguity for downstream use.
\end{itemize}

\paragraph{Dataset Sources} All imagery in the dataset is sourced from OpenAerialMap (OAM, part of the Open Imagery Network / OIN).

\paragraph{Dataset License}  OIN declares that all imagery contained within is licensed as CC BY 4.0 (\url{https://github.com/openimagerynetwork/oin-register}) however some images are labelled as CC BY-NC 4.0 or CC BY-SA 4.0 in their metadata.

To ensure that image providers' rights are upheld, we split these images into license-specific repositories, allowing users to pick which combinations of compatible licenses are appropriate for their application. We have initially released model variants that are trained on CC BY 4.0 + CC BY-NC 4.0 imagery. CC BY-SA 4.0 imagery was removed from the training split, but it can be used for evaluation.

\subsection*{Uses} We anticipate that most users of the dataset wish to map tree cover in aerial orthomosaics, either captured by drones/unmanned aerial vehicles (UAVs) or from aerial surveys such as those provided by governmental organisations.

\paragraph{Direct Use} The dataset supports applications where the user provides an RGB input image and expects a tree (canopy) map as an output. Depending on the type of trained model, the result could be a binary segmentation mask or a list of detected trees/groups of tree instances. The dataset can also be combined with other license-compatible data sources to train models, aside from our baseline releases. The dataset can also act as a benchmark for other tree detection models; we specify a test split which users can evaluate against, but currently there is no formal infrastructure or a leader board for this.

\paragraph{Out-of-Scope Use} The dataset does not contain detailed annotations for trees that are in closed canopy i.e. are touching. Thus the current release is not suitable for training models to delineate individual trees in closed canopy forest. The dataset contains images at a fixed resolution of 10 cm/px. Models trained on this dataset at nominal resolution may under-perform if applied to images with significantly different resolutions (e.g. satellite imagery).

The dataset does not directly support applications related to carbon sequestration measurement (e.g. carbon credit verification) or above ground biomass estimation as it does not contain any structural or species information which is required for accurate allometric calculations~\cite{reierson2021}. Similarly models trained on the dataset should not be used for any decision-making or policy applications without further validation on appropriate data, particularly if being tested in locations that are under-represented in the dataset. See Table~\ref{tab:biome_stats} in this document.

\begin{figure}
	\centering
	\includegraphics[width=\textwidth]{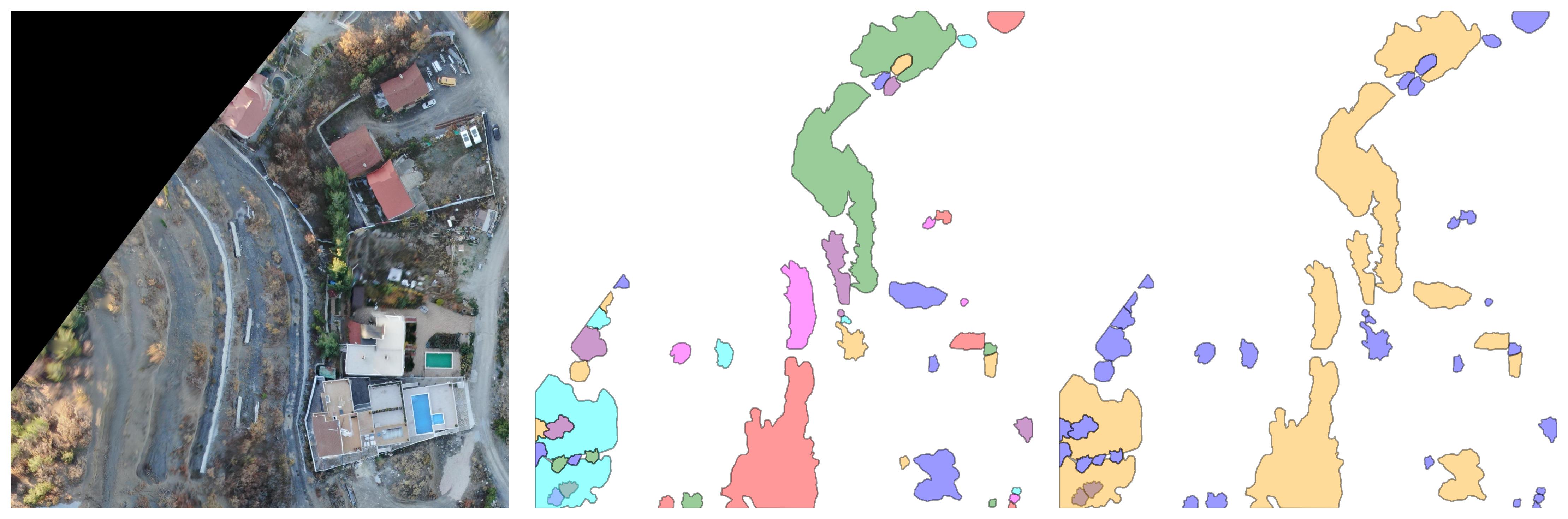}
	\includegraphics[width=\textwidth]{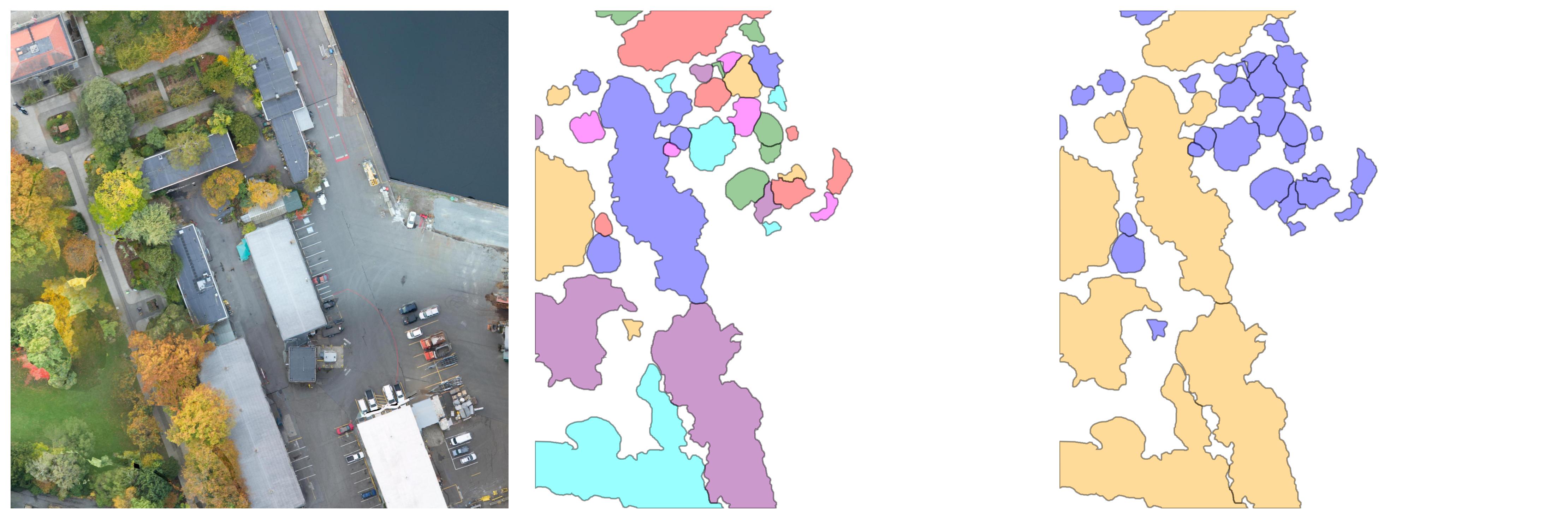}
	\includegraphics[width=\textwidth]{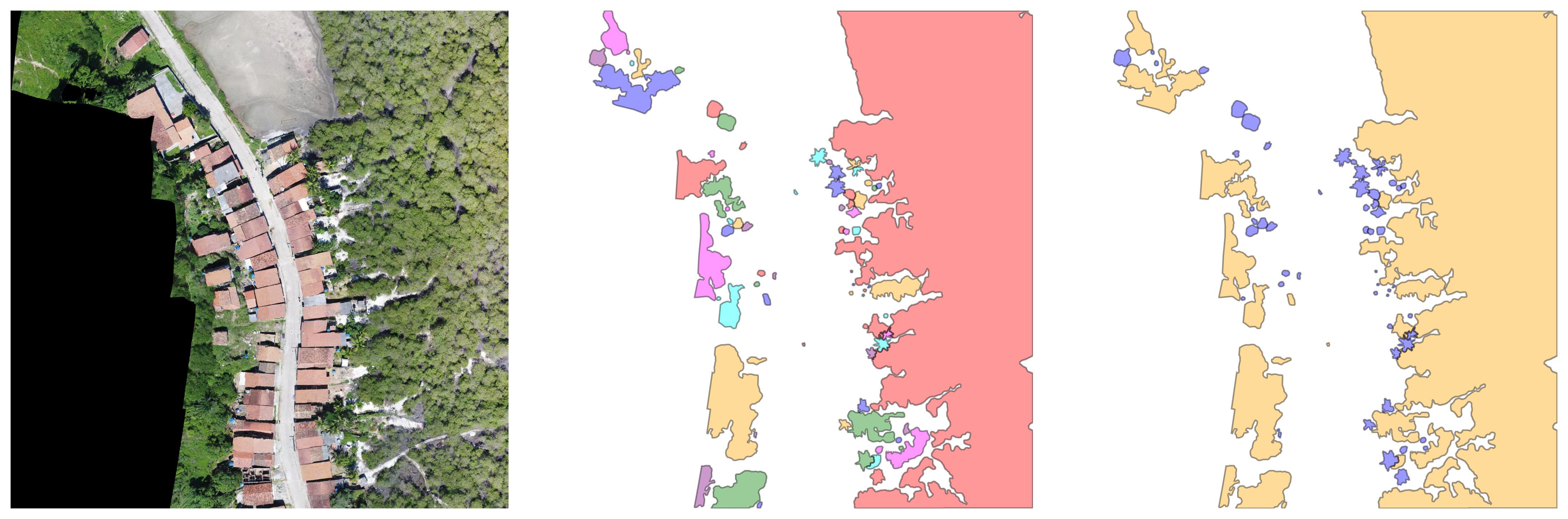}
	\includegraphics[width=\textwidth]{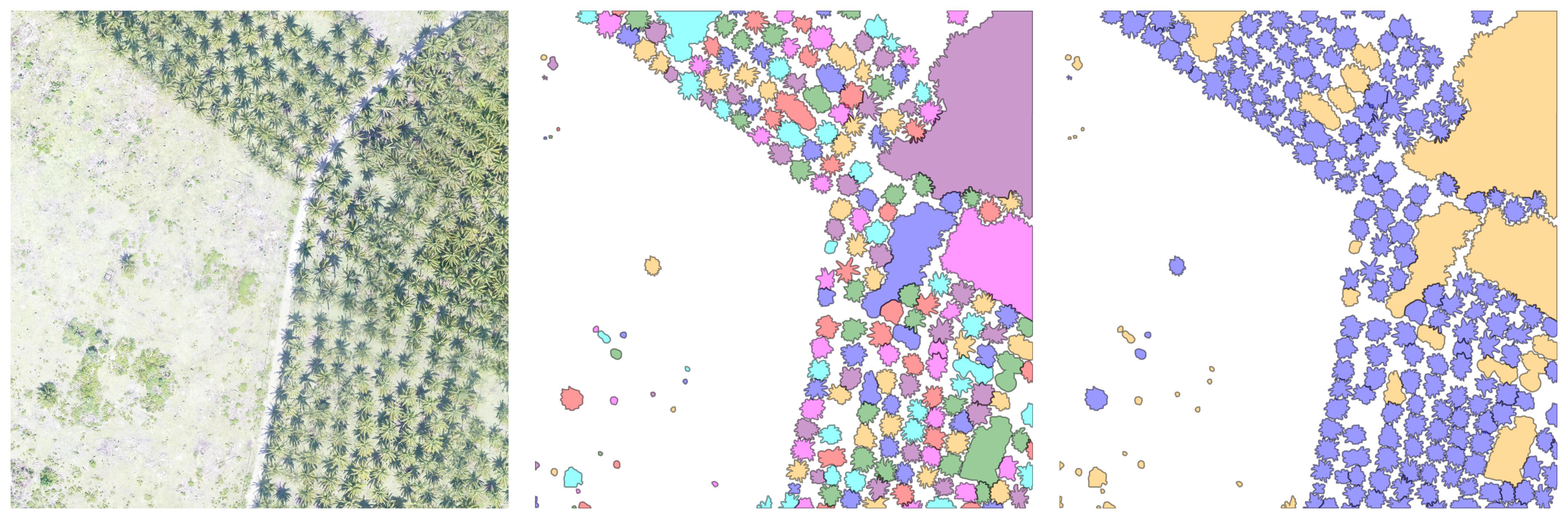}
	\caption{Further example annotations from the OAM-TCD test split. Left: RGB image, Middle: ground truth segmentation randomly coloured by segment ID, Right: coloured by class - blue = tree, orange = tree canopy. All images licensed CC BY 4.0, contributors to Open Imagery Network, top-bottom OAM-TCD IDs: 555,1445,1594,2242.}
	\label{fig:annotation_examples}
\end{figure}

\paragraph{Dataset Structure} The dataset contains pairs of images, semantic masks and object segments (instance polygons). The masks contain instance-level annotations for (1) individual trees and (2) groups of trees, which we label \texttt{canopy}. For training our models we binarise the masks. Metadata from OAM for each image is provided and described in Section~\ref{metadata}. Example annotations are shown in Figure~\ref{fig:annotation_examples}.

The dataset is released with suggested training and test splits, stratified by biome. These splits were used to derive results presented in the main paper. Where known, each image is also tagged with its terrestrial biome index [-1, 14]. This relationship was defined by looking for intersections between tile polygons and reference biome polygons, an index of -1 means a biome wasn't able to be matched. Tiles sourced from a given OAM image are isolated to a single fold (and split) to avoid train/test leakage.

k-fold cross-validation indices within the training set are also provided ($k = 5$). That is, each image is assigned an integer [0, 4] which refers to a validation fold. Users are also free to pick their own validation protocol (for example one could split the data into biome folds), but results may not be directly comparable with results from the release paper.

\subsection*{Dataset Creation}

\paragraph{Curation Rationale} The use-case within Restor~\cite{CROWTHER2022476} is to feed into a broader framework for restoration site assessment. Many users of the Restor platform are stakeholders in restoration projects; some have access to tools like UAVs and are interested in providing data for site monitoring. Our goal was to facilitate training tree canopy detection models that would work robustly in any location. The dataset was curated with this diversity challenge in mind - it contains images from around the world and (by serendipity) covers most terrestrial biome classes. 

It was important during the curation process that the data sources be open-access and so we selected OpenAerialMap as our image source. OAM contains a large amount of permissively licensed global imagery at high resolution (chosen to be  < 10 cm/px for our application).

\subsection*{Source Data}
\paragraph{Data Collection and Processing} We used the OAM API to download a list of surveys on the platform. Using the metadata, we discarded surveys that had a ground sample distance of greater than 10 cm/px (for example satellite imagery). The remaining sites were binned into 1 degree square regions across the world. There are sites in OAM that have been uploaded as multiple assets, and naive random sampling would tend to pick several from the same location. We then sampled sites from each bin and random non-empty tiles from each site until we had reached around 5000 tiles. This was arbitrarily constrained by our estimated annotation budget.

\par Interestingly we did not make any attempt to filter for images containing trees, but in practice there are few negative images in the dataset. Similarly we did not try to filter for images captured in a particular season, so there are trees without leaves in the dataset.

\paragraph{Who are the source data producers?} The images are provided by users of OpenAerialMap / contributors of Open Imagery Network.

\subsection*{Annotation}
\paragraph{Annotation process} Annotation was outsourced to commercial data labelling companies who provided access to teams of professional annotators. We experimented with several labelling providers and compensation strategies.

Annotators were provided with a guideline document that provided examples of how we expected images should be labeled. This document evolved over the course of the project as we encountered edge cases and questions from annotation teams. As described in the main paper, annotators were instructed to attempt to label open canopy trees individually (i.e. trees that were not touching). If possible, small groups of trees should also be labelled individually and we suggested < 5 trees as an upper bound. Annotators were encouraged to look for cues that indicated whether an object was a tree or not, such as the presence of (relatively long) shadows and crown shyness (inter-crown spacing). Larger groups of trees, or ambiguous regions would be labelled as "canopy". Annotators were provided with full size image tiles (2048 x 2048) and most images were annotated by a single person from a team of several annotators.

There are numerous structures for annotator compensation - for example, paying per polygon, paying per image and paying by total annotation time. The images in OAM-TCD are complex and a fixed cost per image was excluded early on as the reported annotation time varied significantly. Anecdotally we found that the most practical compensation structure was to pay for a fixed block of annotation time with regular review meetings with labeling team managers. Overall, the cost per image was between 5-10 USD and the total annotation cost was approximately 25k USD. The labelling companies that we contracted declared that they compensated annotators fairly within their country of domicile. Unfortunately we do not have accurate estimates for time spent annotating all images, but we did advise annotators that if they spent more than 45-60 minutes on a single image that they should flag it for review.

\paragraph{Who are the annotators?} We did not have direct contact with any annotators and their identities were anonymised during communication, for example when providing feedback through managers.

\paragraph{Bias and Risks} There are several potential sources of bias in our dataset. The first is geographic, related to where users of OAM are likely to capture data - accessible locations that are amenable to UAV flights. Some locations and countries place strong restrictions on UAV possession and use, for example. One of the use-cases for OAM is providing traceable imagery for OpenStreetMap which is also likely to bias what sorts of scenes users capture. The sites in OAM-TCD are most likely not restoration projects and we did not find any overlap between sites in OAM and Restor at the time of sampling.

The second is bias from annotators, who were not ecologists. Benchmark results from models trained on the dataset suggest that overall label quality is sufficient for accurate semantic segmentation. However, for instance segmentation annotators had freedom the choose whether to individually label trees or not. This naturally resulted in some inconsistency between what annotators determined was a tree, and at what point to annotate a group of trees as a group. In the main paper, we discuss the issue of conflicting definitions for "tree" among researchers and monitoring protocols.

Figure~\ref{fig:annotation_examples} highlights some of the inconsistencies described above. Some annotators labeled individual trees within group labels; in the bottom plot most palm trees are individually segmented, but some groups are not. A goal for the project is to attempt to improve label consistency, identify incorrect labels and attempt to split group labels into individuals. After annotation was complete, we contracted two different labelling organisations to review (and re-label) subsets of the data; we have not released this data yet, but plan to in the future.

Biases related to biome coverage are discussed in section~\ref{sec:biome_stats} and in Table~\ref{tab:biome_stats}.

The greatest risk that we foresee om releasing this dataset is usage in out-of-scope scenarios. For example, using trained models on imagery from regions/biomes that the dataset is not representative of without additional validation. Similarly there is a risk that users apply the model in inappropriate ways, such as measuring canopy cover on imagery taken during periods of abscission (when trees lose leaves). It is important that users carefully consider timing (seasonality) when comparing time-series predictions.

While we believe that the risk of malicious or unethical use is low - given that other global tree maps exist and are readily available - it is possible that models trained on the dataset could be used to identify areas of tree cover for illegal logging or other forms of land exploitation. Given that our models can segment tree cover at high resolution, it could also be used for automated surveillance or military mapping purposes.

\paragraph{Personally identifiable information} Contact information is present in the metadata for imagery. We do not distribute this data directly, but each image tile is accompanied by a URL pointing to a JSON document on OpenAerialMap where it is publicly available. Otherwise, the imagery is provided at a low enough resolution that it is not possible to identify individual people.

The image tiles in the dataset contain geospatial information which is not obfuscated, however as one of the purposes of OpenAerialMap is humanitarian mapping (e.g. tracing objects for inclusion in OpenStreetMap), accurate location information is required and uploaders are informed that this information would be available to other users. We also assume that image providers had the right to capture imagery where they did, including following local regulations that govern UAV activity. 

An argument for retaining accurate geospatial information is that annotations can be verified against independent sources, for example global land cover maps. This allows for combination with other geo-referenced datasets like multispectral satellite imagery or products like Global Ecosystem Dynamics Investigation (GEDI) data~\cite{Dubayah_2020}.

Contact information is provided in our repository that discusses how users/rights holders can request for imagery to be removed.

\subsection{Image Metadata}
\label{metadata}
We include metadata for the source image of each tile. Some key fields are replicated in the dataset, the full metadata is provided as a URL. The metadata follows the Open Imagery Network specification, described at \url{https://github.com/openimagerynetwork/oin-metadata-spec}. Briefly, this includes the bounding coordinates of the image, its coordinate reference system, ground sample distance, contact information for the image supplier and information about the acquisition (dates, aerial platform, etc.).

\subsection{General Dataset Statistics}
\label{sec:biome_stats}
\par The dataset contains 5072 image tiles sourced from OpenAerialMap; of these 4608 are licensed as CC BY 4.0, 272 are licensed as CC BY-NC 4.0 and 192 are licensed as CC BY-SA 4.0. As described earlier, we split these images into separate repositories to keep licensing distinct. Only around 5\% of imagery in the training split has a less permissive non-commercial license and we are re-training models on only the CC BY portion of the data to maximise accessibility and re-use.

\par The training dataset split contains 4406 images and the test split contains 666 images. All images are the same size (2048x2048 px) and the same ground sample distance (10 cm/px). The geographic distribution of the dataset is shown in the main paper.

\paragraph{Biome Distribution} Table~\ref{tab:biome_stats} shows the number of tiles that correspond to each of the 14 terrestrial biomes described by Olson et. al~\cite{ecoregions}. The majority of the dataset covers (1) tropical and temperate broadleaf forest. Some biomes are clearly under-represented - notably (6) boreal forest/taiga; (9) flooded grasslands and savannas; (11) tundra; and (14) mangrove. Some of these biomes, mangrove in particular, are likely under-represented due to our sampling method (by binned location), as their geographic extent is relatively small. These statistics could be used to guide subsequent data collection in a more targeted fashion.

It is important to note that the biome classification is purely spatial and without inspecting images individually, one cannot make assumptions about what type of landscape was actually imaged, or if it is a natural ecosystem representative of that biome. We do not currently annotate images with a land use category, but this would potentially be a useful secondary measure of diversity in the dataset.

\paragraph{Dataset splits} Since the dataset is relatively small - just over 5000 images - we opted to perform a 5-fold cross validation to better estimate model performance and to allow for training on more data at release time. Folds are stratified over terrestrial biomes using the \texttt{model\_selection.StratifiedKFold} function in \texttt{scikit-learn}~\cite{scikit-learn}. Table~\ref{tab:biome_stats} also shows statistics for the cross-validation folds (e.g. fold size). Since we stratify at the level of source imagery index (\texttt{oam\_id})to avoid leaking tiles between folds, there is some variation in fold size due to differing numbers of tiles from each source image.

\par In addition to cross-validation, approximately 10\% of the images (plus all CC BY-SA 4.0 images) are reserved as a test/holdout set which was not used during hyperparameter tuning and model experimentation. This set is only used to evaluate our final "release" models.

\par As described in the dataset card, validation fold IDs are assigned to each image and filter operations can be used on the dataset table to construct appropriate subsets for training.

\begin{table}
	\centering\
	\begin{tabularx}{\linewidth}{@{} >{\bfseries}Xcccccccc@{} }
		\toprule
		Cross-validation fold&1&  2&  3&  4&  5& Holdout &Total\\
		\cmidrule(lr){1-1}
		Biome&&  &  &  &  &  &\\
		\midrule
		Unmatched&46&  18&  34&  39&  20& 59 & 206\\
		(1) Tropical \& Subtropical Moist Broadleaf Forests &268&  412&  378&  377&  295 & 276 & 2006 \\
		(2) Tropical \& Subtropical Dry Broadleaf Forests&53&  68&  12&  48&  46& 13 & 240\\
		(3) Tropical \& Subtropical Coniferous Forests &11&  6&  13&  19&  10&  6 & 65\\
		(4) Temperate Broadleaf \& Mixed Forests &354&  368&  260&  371&  277&  202 & 1832\\
		(5) Temperate Coniferous Forests &74&  43&  51&  46&  28&  55& 297\\
		(6) Boreal Forests/Taiga&0&  0&  0&  8&  0&  3&11\\
		(7) Tropical \& Subtropical Grasslands, Savannas, and Shrublands&15&  34&  20&  28&  37&  8& 142\\
		(8) Temperate Grasslands, Savannas, \& Shrublands & 4& 5& 14& 15& 12& 34& 84\\
		(9) Flooded Grasslands \& Savannas & 0& 10& 0& 0& 0& 0& 10\\
		(10) Montane Grasslands \& Shrublands & 0& 6& 12& 0& 0& 0& 18\\
		(11) Tundra & 0& 9& 0& 0& 0& 0& 9\\
		(12) Mediterranean Forests, Woodlands, \& Scrub & 12& 11& 27& 16& 12& 2 &80\\
		(13) Deserts \& Xeric Shrublands & 1& 13& 6& 1& 6& 7&34\\
		(14) Mangrove & 12& 0& 2& 1& 12& 1&28\\
		\midrule
		Validation image tiles& 850& 1003& 829& 969& 755& 666& 5072\\
		\bottomrule
	\end{tabularx}
	\caption{Number of image tiles in OAM-TCD for each terrestrial biome class, for each cross-validation fold in the dataset. ``Unmatched" tiles were unable to be matched to a biome via polygon intersection. Also shown is the number of tiles per fold, and the distribution of biomes throughout the entire dataset.}
	\label{tab:biome_stats}
\end{table}

\subsection{Hosting and Access}
\par Our dataset is hosted on two platforms for better availability and to mitigate against host failures:

\begin{itemize}
	\item DOI: \texttt{10.5281/zenodo.11617167}
	\item HuggingFace Hub: \url{https://huggingface.co/datasets/restor/tcd}
	\item Zenodo: \url{https://zenodo.org/records/11617167}
\end{itemize}

\par The release on HuggingFace is provided in Apache Parquet format and can be downloaded using the HuggingFace \texttt{datasets} library. The repository contains images, masks and metadata - including MS-COCO annotation records for each image. The annotation format is technically MS-COCO panoptic (mask images contain RGB-encoded instance IDs), though we do not label all pixels. Labels can be used with many of the segmentation models on the \texttt{transformers} platform with minor modification.

\par Many existing object and instance detection frameworks support MS-COCO format annotations, for example Detectron2 (which we use) and the mmdetection\cite{mmdetection} ecosystem. To more easily support these frameworks, and also to provide an alternative hosting platform, the dataset on Zenodo is provided in MS-COCO format.

\par We provide code in our repository (\texttt{tools/generate\_dataset.py}) to generate MS-COCO annotation files for each split and fold. This tool also supports combining datasets. This script was used to generate the release files uploaded to Zenodo.

\par In the near future we aim to provide a Spatio-Temporal Asset Catalog (STAC) with the dataset to allow standardised querying and access.

\subsection{Croissant}

\par The dataset, as hosted on HuggingFace, contains an automatically generated Croissant metadata record: \url{https://huggingface.co/api/datasets/restor/tcd/croissant}.

\section{Models}

We release the following models on HuggingFace. SegFormer models can be downloaded directly using the 
\texttt{transformers} library, other model weights must be used in conjunction with other libraries or are automatically downloaded with our inference pipeline.

\begin{itemize}
	\item SegFormer mit-b0: \texttt{restor/tcd-segformer-mit-b0}
	\item SegFormer mit-b1: \texttt{restor/tcd-segformer-mit-b1}
	\item SegFormer mit-b2: \texttt{restor/tcd-segformer-mit-b2}
	\item SegFormer mit-b3: \texttt{restor/tcd-segformer-mit-b3}
	\item SegFormer mit-b4: \texttt{restor/tcd-segformer-mit-b4}
	\item SegFormer mit-b5: \texttt{restor/tcd-segformer-mit-b5}
	\item UNet ResNet34: \texttt{restor/tcd-unet-r34}
	\item UNet ResNet50: \texttt{restor/tcd-unet-r50}
	\item Mask-RCNN ResNet50: \texttt{restor/mask-rcnn-r50}
\end{itemize}

\subsection{Model Card(s)}

This model card is reproduced from our HuggingFace repository, using the provided template as reference for headings. Minor changes have been made for readability and formatting, and for the sake of brevity we have combined cards for instance and segmentation models, highlighting differences where appropriate. There is some duplication of information between model and dataset cards.

\begin{itemize}
	\item Developed by: Restor and ETH Zurich
	\item Funding:  Restor and ETH Zurich, supported by a Google.org AI for Social Good grant (ID: TF2012-096892, AI and ML for advancing the monitoring of Forest Restoration)
	\item License: Mask-RCNN and UNet weights are currently licensed as CC BY-NC 4.0. SegFormer weights are licensed under the NVIDIA Source Code License for SegFormer\footnote{See \url{https://github.com/NVlabs/SegFormer/blob/master/LICENSE}, accessed 1 June 2024.}.
\end{itemize}

\paragraph{Semantic Segmentation} models were trained on global aerial imagery from OAM-TCD and are able to accurately delineate tree cover in similar images. The models do not detect individual trees, but provide a per-pixel binary classification (tree/not tree). Post-processing such as connected components could be used to convert results into instance polygons.

\paragraph{Instance Segmentation} models can predict individual trees in open canopy environments. Model output is instance polygons for trees and groups of trees.

\paragraph{Model Sources} References for each model type may be found here:~\cite{he2018mask,ronneberger2015u,xie2021segformer}. Training code for all models is provided in our repository: \url{https://github.com/restor-foundation/tcd} and more information may be found in the accompanying main paper.

\subsection*{Uses}
The primary use-case for these model is assessing canopy cover from aerial images (i.e. percentage of study area that is covered by tree canopy).

\paragraph{Direct Use} Models on their own are suitable for inference on a single image tile (typically 2048 px or smaller). For performing predictions on large orthomosaics, a higher level framework is required to manage tiling source imagery and stitching predictions. Our repository provides a comprehensive reference implementation of such a pipeline and has been tested on extremely large images (country-scale).

The model returns predictions for an entire image. In most cases users will want to predict cover for a specific region of the image, for example a study plot or some other geographic boundary. Some kind of region-of-interest analysis on the results is therefore required. Our linked pipeline repository supports standard shapefile-based region analysis and evaluation.

Direct model usage with the \texttt{transformers} library is straightforward for SegFormers:

\lstset{
	basicstyle=\small\ttfamily,
	frame=tb,
	basewidth=4.5pt, %
	breaklines,
	linewidth=\linewidth %
}

\begin{lstlisting}[language=Python]
	from transformers import AutoImageProcessor
	from transformers import AutoModelForSemanticSegmentation
	import torch
	
	processor = AutoImageProcessor.from_pretrained("restor/tcd-segformer-mit-b0")
	model = AutoModelForSemanticSegmentation.from_pretrained("restor/tcd-segformer-mit-b0")
	
	x = torch.randint(255, (3,512,512))
	inputs = processor(images=[x], return_tensors="pt")
	preds = model(pixel_values=inputs.pixel_values)
\end{lstlisting}

\paragraph{Out-of-Scope Use}

While we trained models on globally diverse imagery, some ecological biomes are under-represented in the training dataset. We therefore encourage users to experiment with their own imagery before using the model for mission-critical use.

Models were trained on imagery at a resolution of 10 cm/px and predictions on other resolutions may be unreliable. We recommend fine-tuning on images with alternative resolutions if that is required.

The model does not predict biomass, canopy height or other derived information. It only predicts the likelihood that some pixel (or polygon) is a tree (canopy). As-is, the model is not suitable for carbon estimation or similar activities.

\subsection*{Bias, Risks, and Limitations}

The main limitation of these models is false positive predictions over objects that look like, or could be confused as, trees. For example large bushes, shrubs or ground cover that looks like tree canopy. This is particularly a concern if images with resolutions significantly different to 10 cm/px. Some conservation/monitoring protocols specify height bounds to determine which objects are considered trees or not; our model does not predict this, but it could be used in conjunction with a LIDAR or photogrammetry surface model to filter appropriate points.

The dataset used to train this model was annotated by non-experts. We believe that this is a reasonable trade-off given the size of the dataset and the results on independent test data, as well as empirical evaluation during experimental use at Restor on partner data. However, there are almost certainly incorrect or inconsistent labels in the dataset and this may translate into incorrect predictions or other biases in model output. We have observed that semantic segmentation models tend to "disagree" with training data in a way that is reasonable (i.e. the aggregate statistics of the labels are good) and we are working to re-evaluate all training data to remove spurious labels. As we note in the main paper, there is no uniform definition for what a tree is, so it is impossible to construct a dataset that satisfies all use-cases unless it also contains per-instance species annotations.

We provide cross-validation results to give a robust estimate of prediction performance, as well as results on independent imagery (i.e. images the model has never seen) so users can make their own assessments. We do not provide any guarantees of accuracy and users should perform their own independent testing prior to deployment.

\subsection*{Training Data} Models were trained using fine-tuning/transfer learning from ImageNet and MS-COCO. Otherwise the only training data used was OAM-TCD. See section~\ref{dataset}.

\subsection*{Training Procedure} We used a 5-fold cross-validation process to tune hyperparameters, before training on the whole training set and evaluating on the OAM-TCD holdout set. The model checkpoints in the \texttt{main} branches of the repository should be considered release versions; this is the default branch used when downloading from HuggingFace Hub.

\paragraph{Semantic Segmentation} Pytorch Lightning was used to train semantic segmentation models.

\paragraph{Instance Segmentation} The Detectron2 framework was used to train Mask-RCNN-based models.

\paragraph{Preprocessing} For SegFormer models, we use the provided pre-processor class that performs normalisation (ImageNet statistics) and converts the input to a Pytorch tensor. We do not resize input images (so that the geospatial scale of the source image is respected) and we assume that normalisation is performed in this processing step and not as a dataset transform. A similar approach was followed to train UNet and Detectron2-based models.

\paragraph{Semantic segmentation training hyperparameters} shown in Table~\ref{tab:semantic_hyper}

\begin{table}
	\centering
	\begin{tabularx}{\linewidth}{@{} >{\bfseries}lX@{} }
		Image size& 1024 x 1024 px\\
		Base learning rate& 1e-4 to 1e-5\\
		Scheduler& Reduce on plateau\\
		Optimizer& AdamW\\
		Batch size& 32\\
		Augmentation& Random crop of source image to 1024x1024, arbitrary rotation, horizontal and vertical flips, colour adjustments\\
		Epochs&75 for cross-validation to ensure convergence; 50 for final models\\
		Normalisation&ImageNet statistics\\
		Loss function&Focal loss (UNet), Cross Entropy (SegFormer)\\
		\bottomrule
	\end{tabularx}
	\caption{Hyperparameters used for semantic segmentation models.}
	\label{tab:semantic_hyper}
\end{table}

\paragraph{Instance segmentation hyperparameters} shown in Table~\ref{tab:instance_hyper}; the training schedule is largely the default suggested in Detectron2, with adjustments to the learning rate, batch size and number of fine-tuning iterations.

\begin{table}
	\centering
	\begin{tabularx}{\linewidth}{@{} >{\bfseries}lX @{} }
		Image size& 1024 x 1024 px\\
		Base learning rate& 1e-3\\
		Schedule& Stepped; Reduced 10x at 80\% and 90\% of training\\
		Optimizer& AdamW\\
		Batch size& 8\\
		Augmentation& Random crop of source image to 1024x1024, arbitrary rotation, horizontal and vertical flips, colour adjustments\\
		Iterations&100,000\\
		Normalisation&ImageNet statistics \\
		Loss function&Mask R-CNN Loss\\
		\bottomrule
	\end{tabularx}
	\caption{Hyperparameters used for instance segmentation (Mask-RCNN) models}
	\label{tab:instance_hyper}
\end{table}

\begin{sidewaysfigure}
	\centering
	\includegraphics[width=\textwidth]{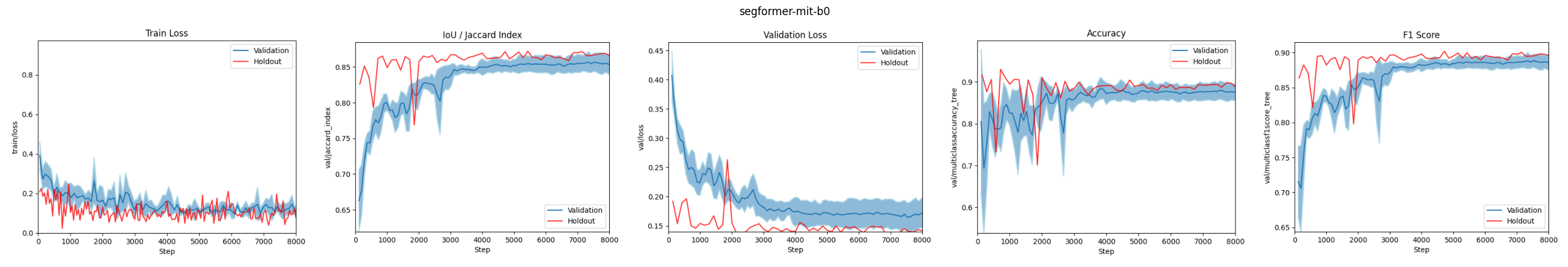}
	\includegraphics[width=\textwidth]{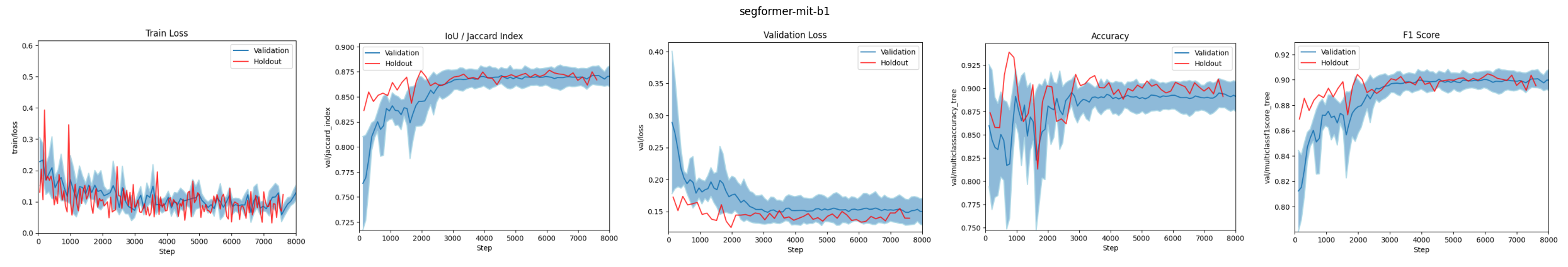}
	\includegraphics[width=\textwidth]{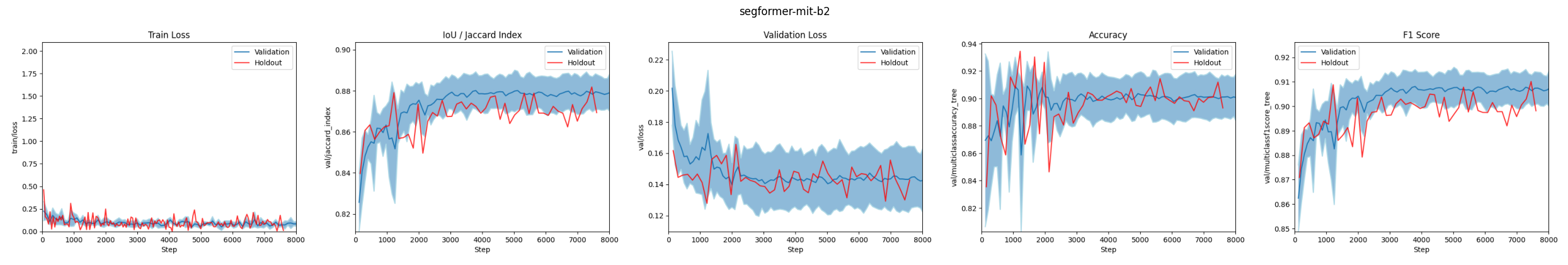}
	\caption{Training/validation curves for SegFormer mit-b0 to mit-b2}
	\label{fig:train_plots_b0_b2}
\end{sidewaysfigure}

\begin{sidewaysfigure}
	\centering
	\includegraphics[width=\textwidth]{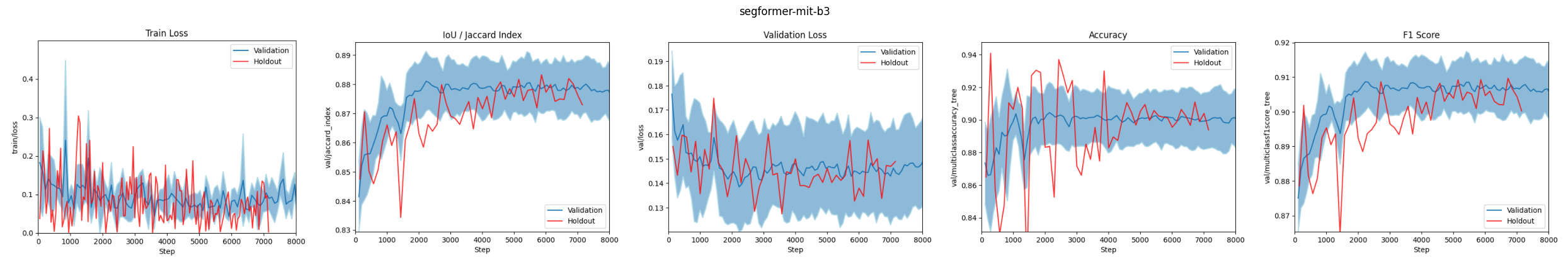}
	\includegraphics[width=\textwidth]{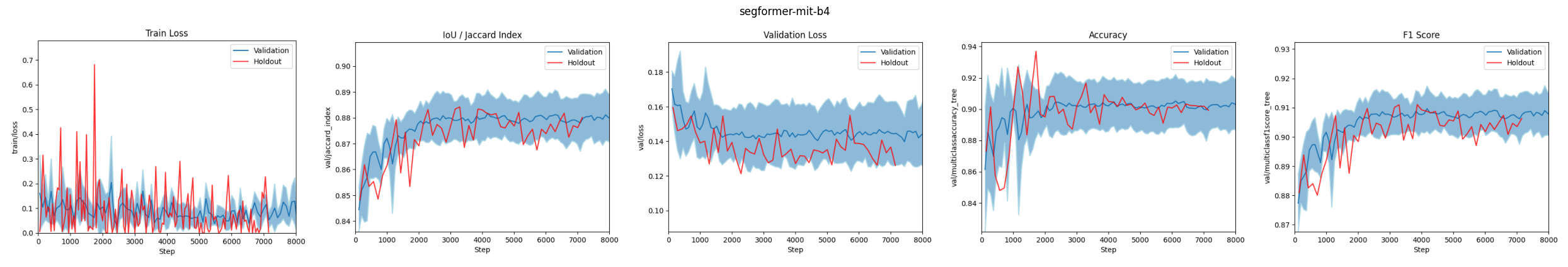}
	\includegraphics[width=\textwidth]{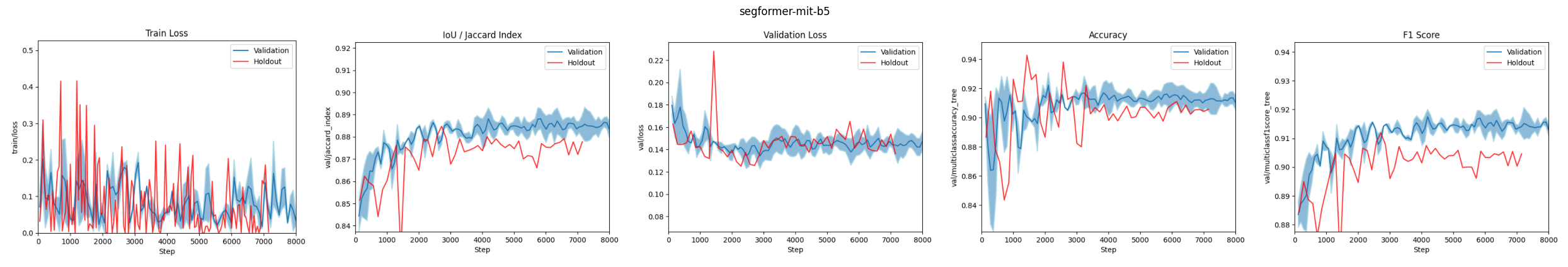}
	\caption{Training/validation curves for SegFormer mit-b3 to mit-b5}
	\label{fig:train_plots_b3_b5}
\end{sidewaysfigure}

\begin{sidewaysfigure}
	\centering
	\includegraphics[width=\textwidth]{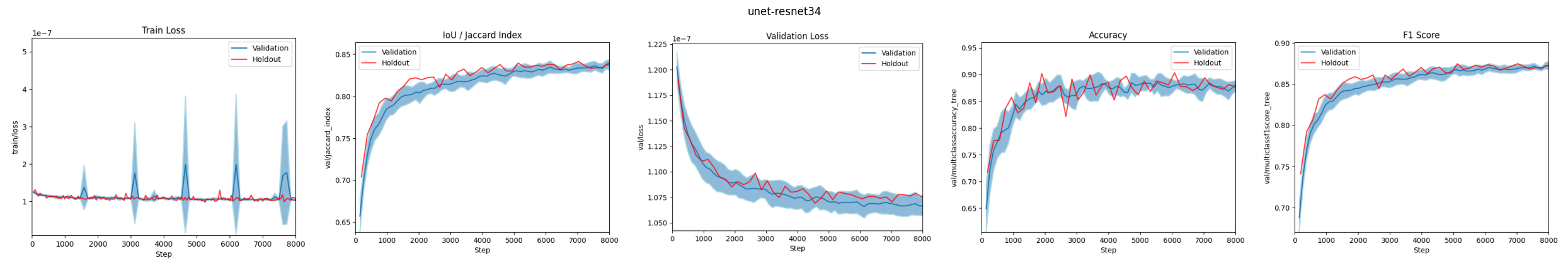}
	\includegraphics[width=\textwidth]{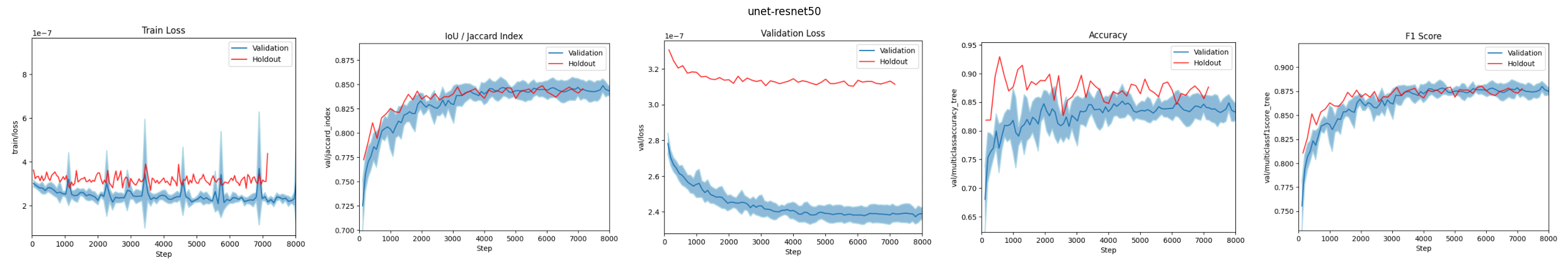}
	\includegraphics[width=0.6\textwidth]{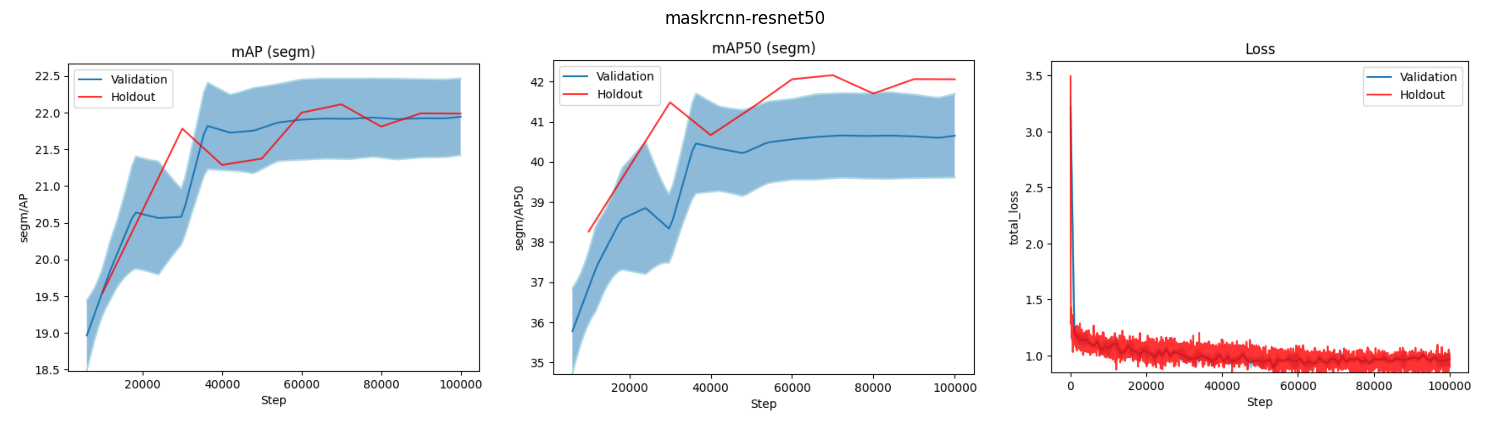}
	\caption{Training/validation curves for UNet models and Mask-RCNN}
	\label{fig:train_plots_unets}
\end{sidewaysfigure}

\paragraph{Training curves} Training curves can be found as Tensorboard records in the model repositories; HuggingFace supports a built-in Tensorboard instance for viewing online. Selected training metrics are shown in Figures~\ref{fig:train_plots_b0_b2},~\ref{fig:train_plots_b3_b5} and ~\ref{fig:train_plots_unets}: aggregated training and validation losses, accuracy, F1 and IoU/Jaccard Index; y-axis labels indicate the metric label used during training for reference against logs. Aggregate curves show the mean and standard deviation bounds for metrics tracked in cross-validation. For Mask-RCNN, we use Detectron2's default logged metrics and report mAP and mAP50 (segmentation). The periodic spikes in validation loss for UNet models appear to be an artifact as overall validation performance is smoothly varying.

\subsection*{Inference and training speed}

Models can be evaluated on CPU, but this requires a large amount of RAM for large tile sizes. In general we find that 1024 px inputs are a sensible upper limit, given the fixed size of the output segmentation masks in SegFormer models (i.e. it is probably better to perform inference in batched mode at 1024x1024 px than try to predict a single 2048x2048 px image).

All models were trained on a single GPU with 24 GB VRAM (NVIDIA RTX3090) attached to a 16-core/32-thread CPU (AMD Ryzen 7950X) with 64GB RAM. All but the largest models can be trained in under a day on a machine of this specification. We provide emissions estimates for trained models in the main paper.

\subsection*{Evaluation}

Models were evaluated on the cross-validation and test splits defined by OAM-TCD. Our pipeline contains evaluation code to compare semantic segmentation output against raster ground truth (e.g. a LIDAR-derived CHM) and instance segmentation output against polygon or keypoint ground truth.

\section{Pipeline}

Our training and prediction pipeline is hosted on GitHub under an Apache 2.0 license: \url{https://github.com/restor-foundation/tcd}. The pipeline supports prediction on large orthomosaics via image tiling. During prediction, images are split into overlapping tiles (by default 1024 px with an overlap of 256 px). Each tile is passed through the model and the results are cached to disk; the pipeline supports batched inference for both instance and semantic segmentation models. End-user documentation for pipeline usage is stored in the repository.

\paragraph{Results caching and output} In order to support predictions over out-of-memory datasets, dataloading and caching during inference are lazy. The pipeline exploits windowed reading and writing supported by GeoTIFFs so that RAM usage is limited to the batch of tiles currently being processed. Results are incrementally stored to a tiled GeoTIFF (for semantic segmentation) or streamed to a shapefile (for instance segmentation). We recommend that virtual rasters (\texttt{vrt} files\footnote{VRT - GDAL Virtual Format, \url{https://gdal.org/drivers/raster/vrt.html}, accessed 1 Jun 2024}) are used for large inputs.

\paragraph{Semantic segmentation post processing} Tile merging follows the approach of~\cite{huang2019tiling}. Half the overlap region for each tile is discarded and the central region is stored to the output cache; we retain predictions at image edges where there is no overlap. We find that SegFormer models have slightly poorer agreement between tile edges, perhaps due the weaker translational invariance in transformer-based models, compared to the convolutional backbone(s) used in UNet + ResNet. 

\paragraph{Instance segmentation post processing} Tree instances that intersect the tile boundary are discarded, but we retain group/canopy predictions. Non-max suppression is performed by default on each tile, separately. After inference is complete, a dissolve operation is performed on all instances to identify overlapping predictions. We then heuristically remove tree polygons that contain multiple centroids from other instances, and then make a final merge decision based on a polygon IoU threshold. Instances are stored in an R-Tree for efficient processing.

\paragraph{Region-based filtering} Results can be optionally filtered via an input geometry, such as a shapefile. This is necessary for reporting results that are constrained to a region of interest, for example measuring canopy cover for a particular plot of land.

\paragraph{Libraries used} Our pipeline relies on several Python packages for geospatial processing which we acknowledge here: \texttt{rasterio}~\cite{gillies_2019} for general image loading and manipulation (which heavily depends on GDAL~\cite{gdal}); \texttt{rtree}~\cite{rtree} for spatially efficient indexing and intersection calculations; shapely for polygon and other geometry processing~\cite{shapely, geos}; \texttt{fiona}~\cite{fiona} for input geometry handling (i.e. Shapefiles, GeoJSON support).

\subsection{Sample Model Predictions}

\begin{sidewaysfigure}
	\centering
	\includegraphics[width=\textwidth]{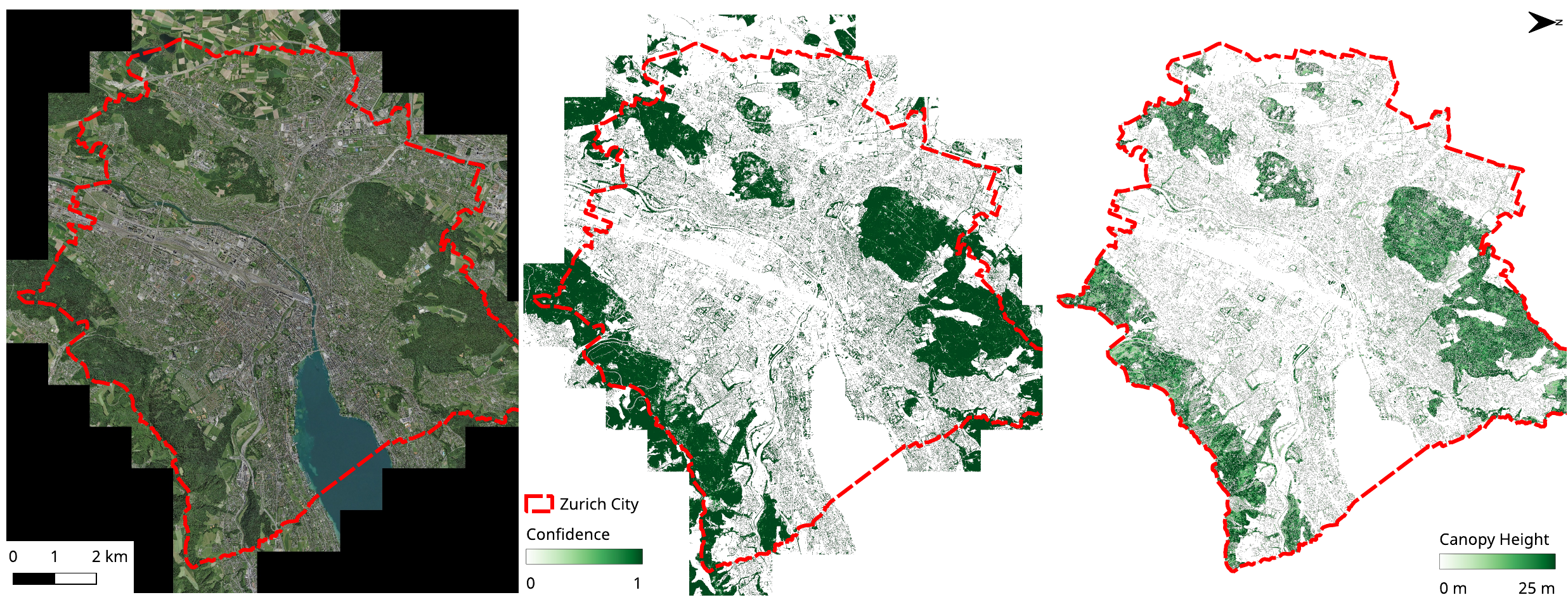}
	\caption{Semantic segmentation predictions for the city of Zurich, with the municipal city boundary marked. Left: 10 cm RGB orthomosaic provided by the Swiss Federal Office of Topography swisstopo/SWISSIMAGE 10 cm (2022), Middle: model predictions, Right: LIDAR CHM (Gruen Stadt Zurich)}
	\label{fig:zurich_semantic}
\end{sidewaysfigure}

\begin{figure}
	\centering
	\includegraphics[width=\textwidth]{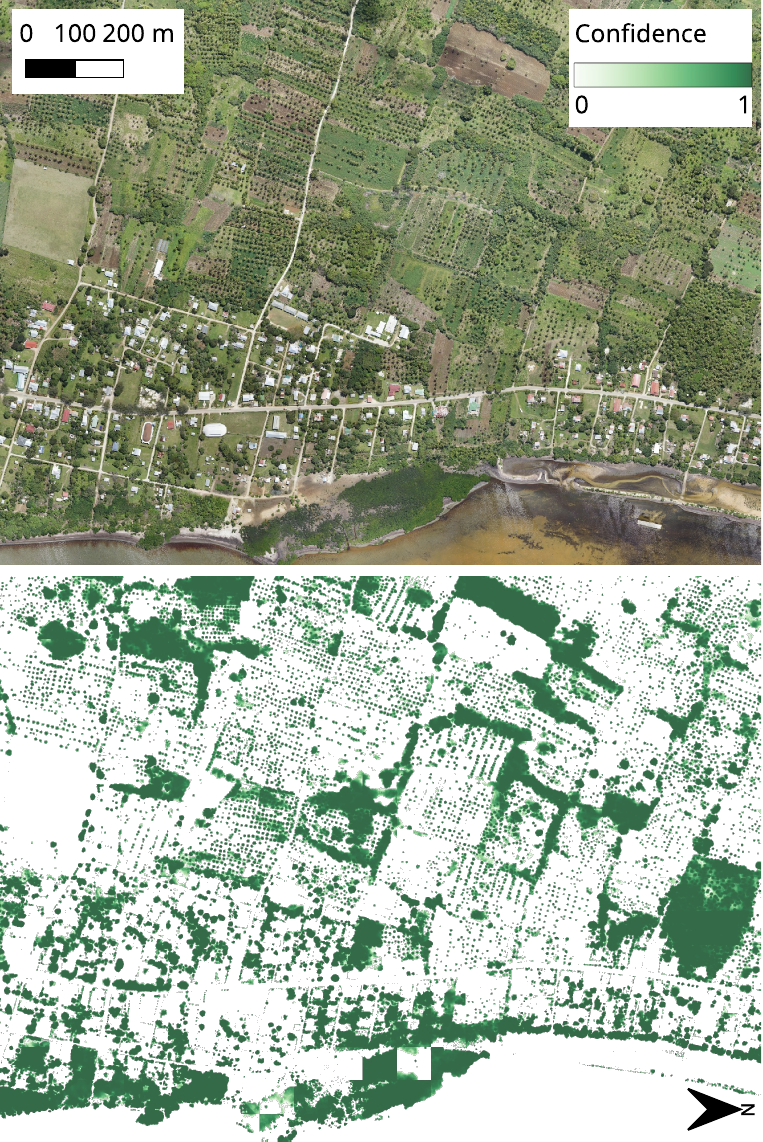}
	\caption{Semantic segmentation predictions for the WeRobotics Open AI challenge image over the Kingdom of Tonga, using the \texttt{restor/tcd-segformer-mit-b5} model. Individual palm trees are clearly segmented. Some uncertain predictions are visible in the lower region of the image near the coast - identifiable as missing/inconsistent tiles.}
	\label{fig:tonga_semantic}
\end{figure}

Figure~\ref{fig:zurich_semantic} and Figure~\ref{fig:tonga_semantic} show complete overviews of model predictions shown in the main paper, using our pipeline.

\par We processed the city of Zurich using our \texttt{segformer-mit-b5} model as a demonstration. Switzerland (Swisstopo) releases imagery over the entire country, but only 1/3 of the cantons are captured each year and the timing of surveys is not consistent (so tree cover is not directly comparable between subsequent captures of the same site). Using a single NVIDIA RTX3090 for compute, the prediction took approximately 1.5 hours for approximately 15k ha of imagery (the image extent is 19.6k ha, but we automatically skip inference on empty tiles).

\section{Further Suggestions for Usage}

\paragraph{Fine-tuning} Since we distribute weights that are compatible with widely used training frameworks, we expect they can be effectively fine-tuned on other tree detection/classification tasks. This approach has been demonstrated with DeepForest~\cite{reierson2021}, for example.

\paragraph{High-resolution panoptic segmentation} Although OAM-TCD only contains labels for tree cover, it would likely be beneficial to augment the dataset with semantic annotations for other features including buildings, roads and geographic features like water cover.

\paragraph{Evaluation of foundational models} There is increasing interest in training foundational models that learn robust representations for geospatial data, given huge repositories of open imagery available. These models can then be used directly (zero-shot) or fine-tuned on downstream tasks. Accordingly, multi-task benchmarks have been proposed (e.g. GEO-Bench~\cite{lacoste2024geo}); OAM-TCD is potentially a useful addition to this landscape due to its flexible label resolution and excellent geographic diversity

\paragraph{Surface map filtering} LIDAR-derived data products are often considered to be a gold standard for forest remote sensing, but classifying point clouds remains a challenging research problem analogous to 2D semantic segmentation. This includes ground extraction, to construct Digital Terrain Models, and classifying points as building, vegetation, etc. In natural environments, once ground has been subtracted, a common assumption is that anything that remains is vegetation. Scans of urban environments must be filtered to discriminate artificial objects as well. High resolution tree canopy maps can be used in conjunction with LIDAR scans to better filter tree-cover compared to point-cloud classification alone.

\paragraph{Unsupervised clustering} Since the dataset contains a large number of tree polygons, it is likely that clustering would be effective to isolate tree crowns with distinct morphology, like palms. As instance segmentation models return masks instead of bounding boxes, predictions are usually less contaminated with background pixels and the mask itself contains geometric information about the detected object. This also makes the outputs more amenable for training species classifiers on prediction crops, if species are known. We provide an example tool for clustering using BioCLIP embeddings in our pipeline.

\end{document}